\documentclass[a4paper,fleqn]{cas-dc}
\usepackage[authoryear]{natbib}

\usepackage{threeparttable}
\usepackage{flushend}
\usepackage{graphicx}

\usepackage{subcaption}
\usepackage{algorithm}
\usepackage{algorithmic}
\usepackage{amsmath}
\usepackage{amssymb}
\usepackage[switch]{lineno}

\def\tsc#1{\csdef{#1}{\textsc{\lowercase{#1}}\xspace}}
\tsc{WGM}
\tsc{QE}

\begin{document}
\begin{sloppypar} 

\let\WriteBookmarks\relax
\def\floatpagepagefraction{1}
\def\textpagefraction{.001}
\let\printorcid\relax

\makeatletter
\renewcommand{\fnum@figure}{Fig. \thefigure.\@gobble}
\makeatother

\shorttitle{}    

\shortauthors{}  

\title [mode = title]{Aerial-ground LiDAR place recognition with patch-level self-supervised learning and expanded reciprocal re-ranking}  


\tnotetext[1]{This research was supported by Jiangxi Provincial Natural Science Foundation (No.20252BAC200598, No.20261BCG330036)}

\affiliation[1]{organization={Department of Geomatics Engineering},
            addressline={University of Calgary}, 
            city={Calgary},
            postcode={T2N 1N4}, 
            country={Canada}}

\affiliation[2]{organization={School of Advanced Manufacturing},
            addressline={Nanchang University}, 
            city={Nanchang},
            postcode={330031}, 
            country={China}}

\affiliation[3]{organization={School of Electrical and Electronic Engineering},
            addressline={Nanyang Technological University}, 
            postcode={639798}, 
            country={Singapore}}
            
\affiliation[4]{organization={School of Remote Sensing and Information Engineering},
            addressline={Wuhan University}, 
            city={Wuhan},
            postcode={430079}, 
            country={China}}

\author[1]{Yandi Yang}

\ead{yandi.yang@ucalgary.ca}

\credit{Conceptualization of this study, Methodology, Software, Validation, Writing – original draft}

\author[2]{Xianghong Zou}
\cormark[1]
\ead{ericxhzou@ncu.edu.cn}
\credit{Conceptualization of this study, Methodology, Writing review}

\author[3]{Jianping Li}
\ead{jianping.li@ntu.edu.sg}
\credit{Methodology, Writing review}

\author[4]{Haofeng Xie}
\ead{xiehaofeng@whu.edu.cn}
\credit{Methodology, Writing review}

\author[1]{Saurav Uprety}
\ead{saurav.uprety1@ucalgary.ca}
\credit{Data collection, Writing review}

\author[1]{Hongzhou Yang}
\ead{honyang@ucalgary.ca}
\credit{Data collection, Writing review}

\author[1]{Naser El-Sheimy}
\ead{elsheimy@ucalgary.ca}
\credit{Conceptualization of this study, Methodology, Writing review, Project administration, Funding acquisition}
\cortext[1]{Corresponding author}



\begin{abstract}
LiDAR place recognition determines one's position on a prior point cloud map. The most studied ground-level LiDAR place recognition suffers from pre-visit requirements, incomplete coverage, and limited perspectives. Using pre-acquired, full-coverage Airborne Laser Scanning (ALS) data as an aerial prior map overcomes these drawbacks, making cross-view place recognition necessary and advantageous. However, aerial-ground LiDAR place recognition faces significant challenges, including the domain gap between aerial and ground point clouds, false positives during initial retrieval, and a lack of large-scale benchmarks. To address these challenges, we present a novel retrieval and re-ranking framework for aerial-ground LiDAR place recognition. Based on the priors that neighboring point cloud patches share similar semantics with anchor patch, our retrieval network introduces patch-level self-supervised learning modules at multiple scales and integrates with scene-level learning to improve global feature discriminativeness between aerial and ground point clouds. Furthermore, leveraging the structured spatial distribution of ALS point clouds, we introduce an Expanded Reciprocal (ER) re-ranking algorithm to exploit neighborhood information maximally and refine each feature based on neighbor features, which are then used to update the similarity matrix for final ranking. To evaluate our approach, we establish a new large-scale benchmark named CS-Urban-Scenes. Collected via a backpack mobile mapping system with post-processed kinematic (PPK) optimized trajectories, the dataset features an 18.1 km trajectory and a 7.2 km$^2$ coverage area. Extensive experiments demonstrate that our retrieval network outperforms existing state-of-the-art (SOTA) methods, achieving a 9.8\% improvement in average Recall@1 and a 3.2\% improvement in average Recall@1\% on the CS-Urban-Scenes, while also showing the best performance on the CS-Campus3D dataset. Additionally, our ER re-ranking algorithm further boosts the average Recall@1 by 4.9\% on CS-Campus3D and 10.2\% on CS-Urban-Scenes without additional training.

\end{abstract}








\maketitle


\section{Introduction}

\begin{figure}[]
\centering
\includegraphics[width=\linewidth, keepaspectratio]{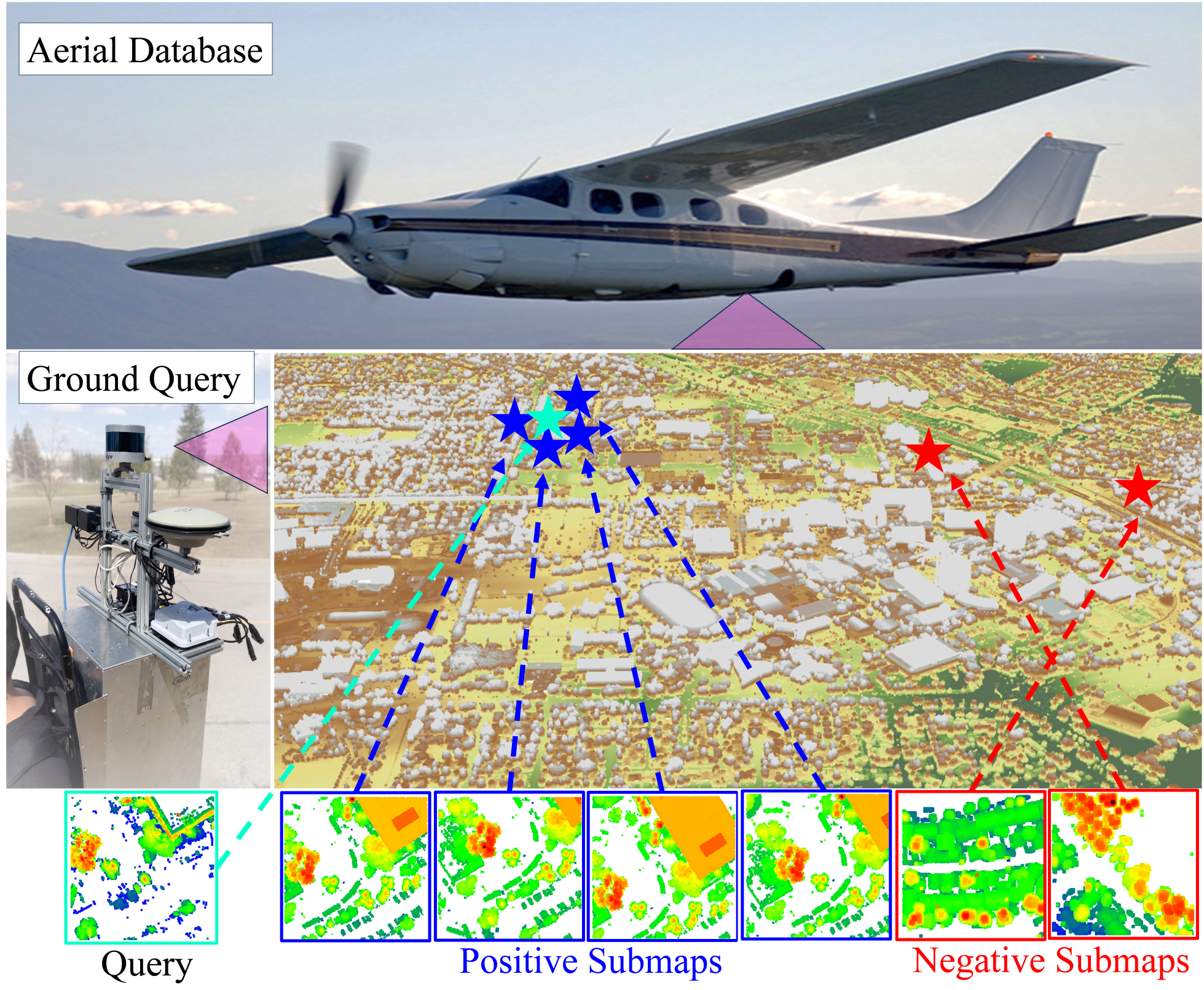}
\caption {Diagram of aerial-ground LiDAR place recognition. Within a prior ALS point cloud database, submap locations are denoted by asterisks: \textcolor{blue}{blue} and \textcolor{red}{red} indicate positive and negative candidates, respectively, relative to the ground query.}
\label{fig_prob}
\end{figure}

\begin{figure*}
\centering
\includegraphics[width=0.95\textwidth, keepaspectratio]{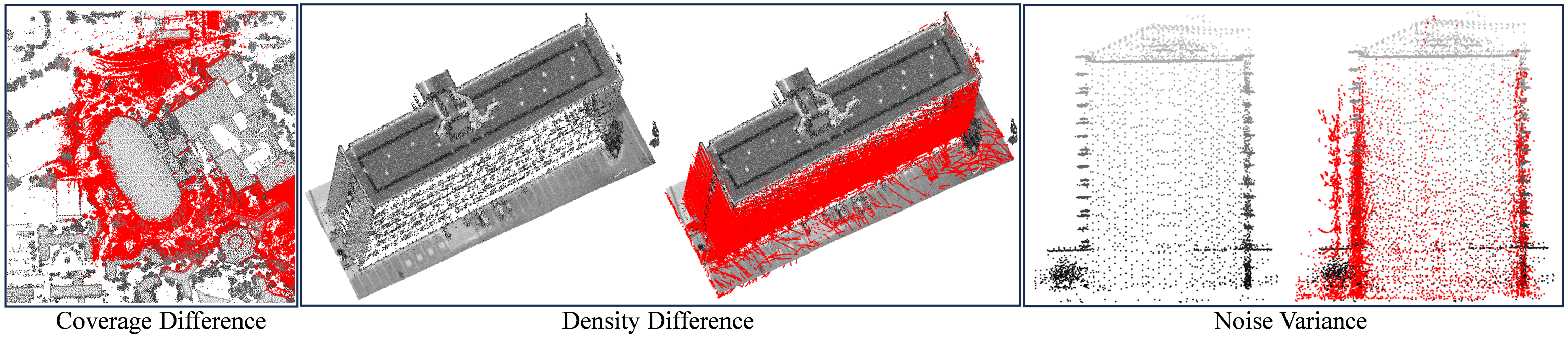}
\caption {Domain gap between aerial and ground point clouds. Gray and red points denote the aerial and ground data, respectively.}
\label{fig_difficulties}
\end{figure*}

Reliable localization is essential for Mobile Mapping Systems (MMS) \citep{schwarz2004mobile} and robotics \citep{li2026aeos} applications. As a core task, place recognition enables an MMS or robot to identify its position within a prior map. Standard solutions leverage Global Navigation Satellite Systems (GNSS), Inertial Navigation Systems (INS), or visual sensors. However, GNSS signals are easily obstructed in urban canyons due to signal occlusion and multipath effects \citep{nassar2006combined}. INS-based odometry suffers from cumulative drift without external corrections \citep{el2020inertial}. Although various visual localization methods have been proposed \citep{jia2026georanker,yang2026agi2p}, they remain sensitive to illumination and viewpoint changes. In contrast, LiDAR directly provides 3D point clouds and is robust to illumination changes. As LiDAR becomes more affordable and easier to integrate onto MMSs, LiDAR-based place recognition has become increasingly important for applications such as SLAM \citep{li2023whu} and global localization \citep{wu2026leader}. A common place recognition pipeline relies on prior street-view point cloud maps \citep{zou2026lifelongpr} or image databases \citep{lee2026lsv}. However, ground-level place recognition remains limited as the ground prior maps depend on pre-visiting the region, cannot fully cover the entire area, and suffer from blockages of ground perspectives.

Airborne Laser Scanning (ALS) point clouds, acquired from aerial platforms, have been made publicly available by many government agencies. Compared with ground point cloud maps, they are pre-collected and geo-referenced, cover complete regions at city or national scales, and provide richer details. These characteristics make aerial-ground LiDAR place recognition practical in mobile mapping applications. Take ground localization as an example. Within areas covered by ALS data, an MMS or robot can determine its position using only ground LiDAR. This approach can circumvent the failures of other sensors, such as GNSS outages in urban canyons, long-term drift in IMUs, and illumination or viewpoint variations in visual sensors. Consequently, this can serve as an external localization source for multi-sensor fusion. The diagram for aerial-ground LiDAR place recognition is detailed in Figure~\ref{fig_prob}. Given a ground query, the task is to retrieve the closest aerial submap from the aerial database.

However, several challenges remain in aerial-ground LiDAR place recognition. (1) Domain gap \citep{zou2023patchaugnet,yang2024ubiquitous} between aerial and ground heterogeneous point clouds: As shown in Figure ~\ref{fig_difficulties}, different data coverages cause only partial overlap and the varied density results in a lack of reliable point correspondences. Furthermore, different sensor noises exist due to the varying scanning distances. These factors pose challenges to the distinctiveness of global features between cross-view data. (2) Numerous false positives in the initial retrieval: Due to repetitive buildings and structure in aerial point cloud database, relying solely on a single global descriptor often causes mismatches \citep{zhong2017re,yang2025cheb}. This one-way search ignores the mutual relationship between the query and other top-ranked candidates in the database. (3) Lack of large-scale benchmarks: Existing aerial-ground datasets are mostly limited to small areas \citep{guan2023crossloc3d}. They lack the long trajectories and environmental diversity to test localization in urban areas. Current methods struggle with these challenges due to two primary reasons. (1) Existing retrieval approaches focus on scene-level metric learning \citep{komorowski2021minkloc3d, vidanapathirana2022logg3d} and ignore feature representation learning at the patch level. Although sufficient for single-platform scenarios, they are vulnerable to the cross-domain gap between aerial and ground point clouds. (2) Re-ranking serves as a post-processing step to refine initial retrieval results. Conventional re-ranking pipelines \citep{sarfraz2018pose, shao2023global} rarely exploit the structured spatial distribution of the database, relying solely on feature distances while ignoring the spatial consistency among neighboring candidates. They are susceptible to similar false positives in large-scale ALS point cloud database, thereby degrading retrieval accuracy.

To address these challenges, we propose a comprehensive aerial-ground LiDAR place recognition framework and establish a new large-scale dataset in dense urban areas. The main contributions of this paper are as follows.

\begin{enumerate}[1)]
\setlength{\leftmargin}{0em}

\item We propose a new LiDAR place recognition pipeline that incorporates patch-level self-supervised learning at multiple scales with scene-level learning to improve global feature representation, thus mitigating the domain gap between aerial and ground point clouds.

\item We present a new Expanded Reciprocal (ER) re-ranking algorithm by considering the structured spatial distribution of the aerial database. Exploiting neighborhood relationships from both the query and database sides, expanding reciprocal neighbors refines the initial global features and mitigates false positives without additional network training.

\item Comprehensive experiments are conducted on the newly proposed CS-Urban-Scenes dataset in challenging urban environments, which features an 18.1 km ground trajectory and a 7.2 $\text{km}^2$ ALS coverage area. The results demonstrate that both our retrieval and re-ranking methods outperform existing SOTA approaches.

\end{enumerate}

The remainder of this paper is organized as follows. Section~\ref{sec_2} reviews the existing literature. Section~\ref{sec_3} introduces the proposed patch-level learning and re-ranking framework, while Section~\ref{sec_4} describes the characteristics of the CS-Urban-Scenes dataset. Extensive quantitative and qualitative evaluations, along with detailed analyses, are presented in Section~\ref{sec_5}. Finally, Section~\ref{sec_6} concludes the paper.

\section{Related work}
\label{sec_2}

\subsection{Patch-level self-supervised learning}
Image self-supervised learning (SSL) increasingly adopts image patches as a processing unit. Although initial methods predict relative positions between patches \citep{doersch2015unsupervised} or concatenate patch features randomly \citep{misra2020self}, vision Transformers (ViTs) have demonstrated the potential for denser image representations. Masked autoencoders \citep{he2022masked} improve this by masking random patches and reconstructing the original image in pixels. DINOv3 \citep{simeoni2025dinov3} introduces Gram anchoring to mitigate the degradation of patch-level consistency. SelfPatch is a visual pretext task for learning better patch-level representations \citep{yun2022patch}. Some methods\citep{wang2021patchmatchnet} also utilize patch-to-patch matching in multi-view stereo. In visual place recognition, CNNs \citep{hausler2021patch} can be used to extract patch-level descriptors. Transformer's self-attention mechanism is leveraged to select a set of key patches, filtering out distractions for efficient re-ranking \citep{zhu2023r2former}.

Several patch-based point cloud learning algorithms have been proposed, such as MLP-based methods \citep{long2022pc2}. Transformer-based methods \citep{gao2025dap} adopt a "mask-and-reconstruct" objective, where an encoder learns latent representations from unmasked patches, while a decoder reconstructs the masked ones based on these features. In addition, Mamba-based point cloud analysis \citep{zha2025pma} uses patch tokens as units. In addition to generating patches using the K-Nearest Neighbors (KNN) and Farthest Point Sampling (FPS) algorithms, Octformer \citep{wang2023octformer} generates patches based on an octree structure, FlatFormer \citep{liu2023flatformer} partitions the point cloud into equal-sized groups, and PTv3 \citep{wu2024point} utilizes space-filling curves to serialize points. PointTPA forms structured local patches from unorder point clouds for better 3D scene understanding \citep{liu2026pointtpa}.

\subsection{LiDAR place recognition}
Point cloud features can be extracted directly by both handcrafted \citep{yuan2024btc,lim2025kiss} and deep learning \citep{komorowski2021minkloc3d,vidanapathirana2022logg3d,zou2023patchaugnet} algorithms, which are then used to search the database to identify revisited areas. Furthermore, some approaches partition raw point clouds into segments before feature extraction. SegMap \citep{dube2020segmap} decomposes point clouds into segments and matches learning-based segment descriptors for retrieval. SSC \citep{li2021ssc} applies semantic information into Scan Context \citep{kim2018scan} for more effective representation. SGPR \citep{kong2020semantic} forms the point clouds into a semantic graph and models place recognition into a graph matching problem. Point clouds can also be projected into range images \citep{luo2025overlapmamba} or Bird’s-Eye-View (BEV) images \citep{skuddis2026bev} to leverage 2D computer vision techniques. Nevertheless, domain gap between aerial and ground data causes segmentation inconsistencies, resulting in mismatched point cloud clusters. Furthermore, projecting point clouds into images yields inconsistent representations. For example, due to the different coverages between aerial and ground data, the projected BEV images vary significantly, which degrades feature learning during network training.

For aerial-ground LiDAR place recognition, CrossLoc3D \citep{guan2023crossloc3d} extracts features through multi-grained voxelization and multi-scale sparse convolution, followed by an iterative refinement process. GAPR \citep{jie2023heterogeneous} employs MinkFPN as the backbone for feature extraction, followed by a transformer encoder to capture the overlaps between ground and aerial point clouds. HOTFormerLoc \citep{griffiths2025hotformerloc} performs hierarchical attention to facilitate multi-scale feature interaction. Both BEV features from CNN and point features from sparse convolution are extracted \citep{wang2025cross} and the experiments are conducted on two datasets, both of which comprise data collected by a UAV and a UGV. However, these approaches only consider scene-level supervision and neglect fine features at the patch level, which makes the generated global descriptors easily corrupted by the domain gap between aerial and ground data. Some methods rely on other publicly available maps, which either suffer from the data imbalance of OpenStreetMap (OSM) \citep{cho2022openstreetmap} or require other sensors for orientation when using satellite images \citep{shi2026l2rsi}. 

\subsection{Re-ranking}
Retrieval results from place recognition can be further refined with a re-ranking module. Local features can be utilized for geometric verification \citep{cao2020unifying} or to estimate a similarity score \citep{xiao2025locore} to refine image retrieval results. Local features can also be utilized in re-ranking point cloud retrieval \citep{vidanapathirana2023spectral}. However, the coverage difference between aerial and ground point clouds limits the extraction of reliable local features across the two sources.

Other methods rely solely on global features for re-ranking. Query expansion refines the query using the top-ranked features \citep{radenovic2018fine,shao2023global}. Contextual information is considered to improve retrieval performance \citep{zhong2017re,sarfraz2018pose}. Global features can also be propagated and refined through a graph \citep{zhang2023graph,yang2025cheb}. However, these re-ranking methods neglect the spatial distribution of ALS point clouds, leading to performance degradation in ground-to-aerial retrieval.

Re-ranking can also be implemented via learnable modules. $R^2\text{Former}$ unifies image retrieval and re-ranking for place recognition using pure transformers \citep{zhu2023r2former}. CVNet utilizes an end-to-end learnable 4D CNN to predict the similarity between two images \citep{lee2024correlation}. Despite their strong performance, these methods require computationally expensive network training. Although Large Language Models (LLMs) have been utilized in image retrieval re-ranking \citep{tharwat2026indexing} and point cloud understanding \citep{xu2025pointllm}, the scarcity of point clouds remain a bottleneck in LiDAR place recognition tasks.

\section{Proposed method}
\label{sec_3}

\begin{figure*}
\centering
\includegraphics[width=0.95\textwidth, keepaspectratio]{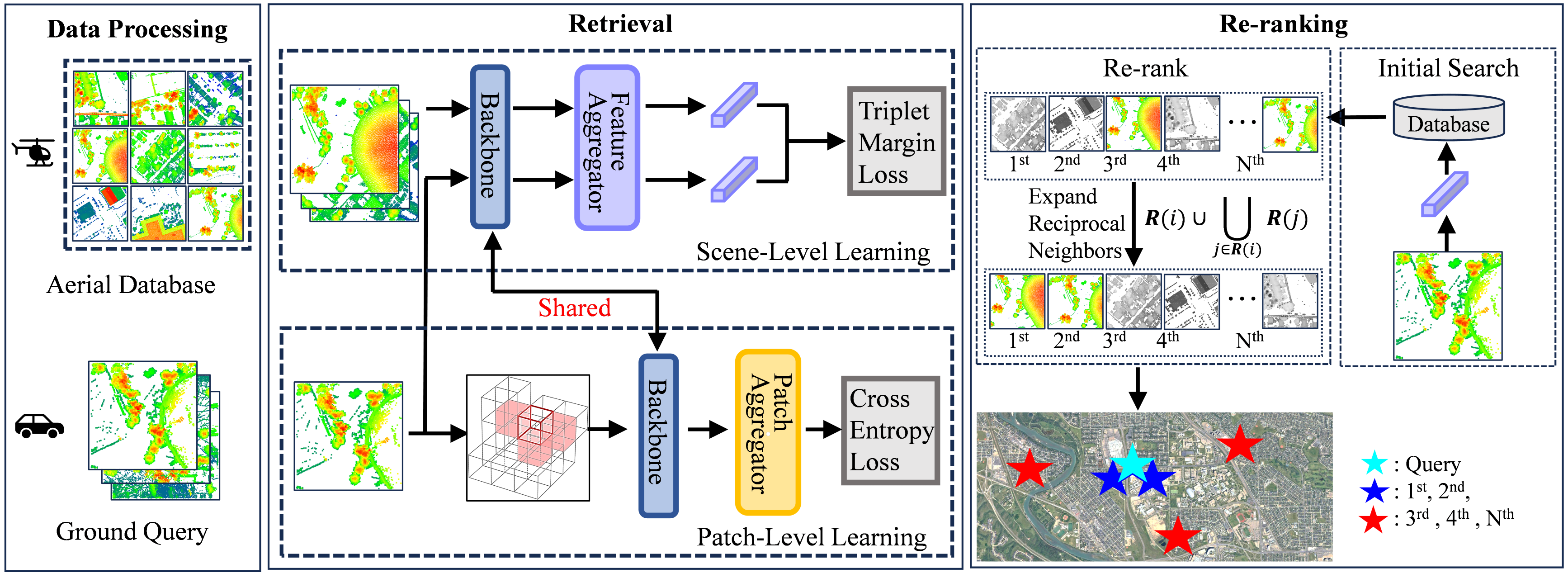}
\caption {Overview of our aerial-ground LiDAR place recognition method. Given a ground query and an aerial point cloud database, we train the network through on both scene and patch levels at the retrieval stage. The scene-level learning is to pull positive pairs closer and push negatives apart, while the multi-scale patch-level learning relies on self-supervised learning. During the re-ranking stage, the initial retrieval results are refined using our Expanded Reciprocal (ER) re-ranking algorithm.}
\label{fig_overview}
\end{figure*}

The proposed framework aims to bridge the cross-source domain gap and filter out numerous false positives from the initial retrieval, thereby enhancing place recognition accuracy between aerial and ground point clouds. Figure~\ref{fig_overview} illustrates the overall workflow of our method, and the problem is defined in ~\ref{sec_3_1}. Self-supervised patch-level learning operates at multiple octree scales, and the network is optimized through a joint loss that combines patch-level with scene-level loss described in Section~\ref{sec_3_2}. Finally, as demonstrated in Section~\ref{sec_3_3}, expanded reciprocal neighbors are leveraged to refine the global features and compute updated pairwise distances, which are then used to re-rank the initial candidate set.

\subsection{Problem formulation}
\label{sec_3_1}

Let $\mathcal{D} = \{M_1, M_2, \dots, M_m\}$ be a database consisting of $m$ aerial LiDAR submaps. Let $P_q $ be a ground query point cloud captured at an unknown location. The proposed framework follows a two-stage pipeline: first retrieval and then re-ranking.

\textbf{Retrieval.} The objective of the initial retrieval stage is to map these heterogeneous point clouds into a shared feature space via an embedding function $f(\cdot) = a(h(\cdot))$, where $h(\cdot)$ extracts multi-scale local features and $a(\cdot)$ aggregates them into a global descriptor. For the query $P_q$, its global descriptor is matched against the database $\mathcal{D}$ using the $L_2$ norm to yield a list of top-$k$ retrieval candidates:
\begin{equation}
    \mathcal{C}_R = \{P_{R_1}, P_{R_2}, \dots, P_{R_k}\},
\end{equation}
where $\mathcal{C}_R$ is sorted in ascending order based on the feature distance $\|f(P_q) - f(P_i)\|_2$. To optimize the embedding function, the network enforces a triplet constraint on the global descriptors:
\begin{equation}
    \|f(P_q) - f(P_p)\|_2 < \|f(P_q) - f(P_n)\|_2,
\end{equation}
where $P_p \in \mathcal{D}$ and $P_n \in \mathcal{D}$ denote the geographically close positive sample and distant negative sample relative to $P_q$, respectively.

\textbf{Re-ranking.} To filter out numerous false positives, a global feature-based re-ranking stage is introduced. The initial global features are refined to compute updated pairwise distances (Section ~\ref{sec_3_3}), which are then used to permute the initial candidate set into the final re-ranked list $\mathcal{C}_{RR}$:
\begin{equation}
    \mathcal{C}_{RR} = \{P_{RR_1}, P_{RR_2}, \dots, P_{RR_k}\}.
\end{equation}
An effective re-ranking algorithm ensures that the top-ranked candidates are geographically closer to the query, satisfying the constraint:
\begin{equation}
    \mathcal{D}_{geo}(\mathbf{x}_q, \mathbf{x}_{RR_i}) \le \mathcal{D}_{geo}(\mathbf{x}_q, \mathbf{x}_{R_i}), \quad \forall i,
\end{equation}
where $\mathbf{x}_q$, $\mathbf{x}_{R_i}$, and $\mathbf{x}_{RR_i}$ denote the ground-truth geolocations of $P_q$, $P_{R_i}$, and $P_{RR_i}$, respectively, and $\mathcal{D}_{geo}(\cdot)$ represents the Euclidean distance of the geographical coordinates. 

\subsection{Retrieval with multi-scale patch-level learning}
\label{sec_3_2}
Considering that neighboring point cloud patches share similar semantics with the anchor patch, we utilize a self-supervised learning scheme to improve global feature representations across aerial and ground heterogeneous data. In this section, we first introduce the backbone and feature aggregator, and then detail the multi-scale patch-level self-supervised learning scheme. Finally, we formulate a training loss that combines self-supervised and metric learning objectives.

\begin{figure*}
\centering
\includegraphics[width=0.95\textwidth, keepaspectratio]{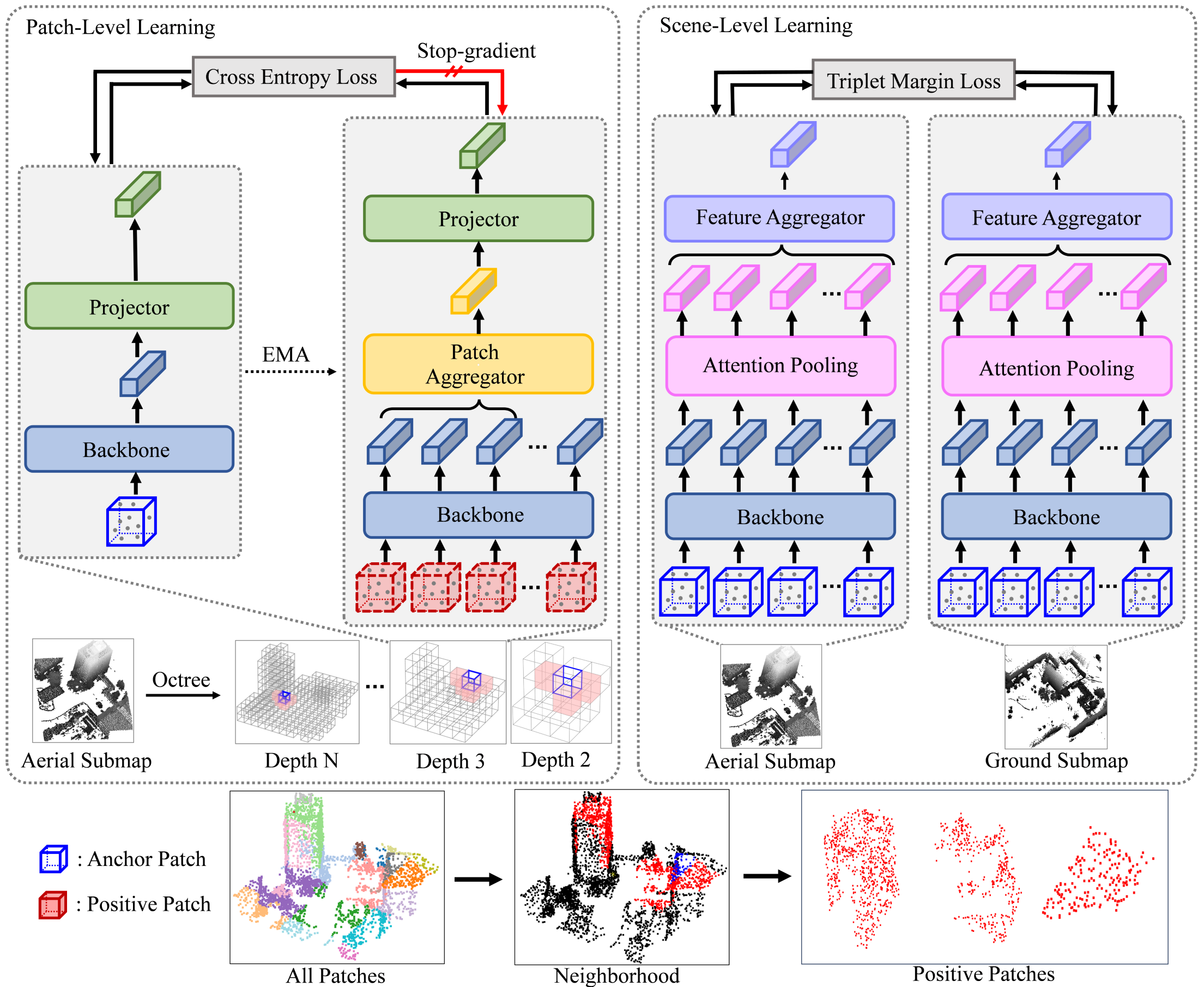}
\caption {Architecture of our retrieval network, which integrates multi-scale patch-level self-supervised learning modules and scene-level metric learning. Our patch-level learning operates under the assumption that neighboring patches and the anchor patch have semantic similarity.}
\label{fig_network}
\end{figure*}

Figure ~\ref{fig_network} shows the architecture of our retrieval network. Scene-level supervision is implemented via metric learning using positive and negative pairs between the ground and aerial data, while patch-level self-supervised learning is performed on independent single point clouds across multiple scales of an octree.

\textbf{Backbone.} We utilize OctFormer \citep{wang2023octformer} as the backbone $h(\cdot)$ to extract features due to its scalability in large-scale point clouds. Given an input point cloud $P$, OctFormer serializes it into an ordered sequence of non-empty octree nodes via Z-order serialization. Each non-empty node at depth $o \in \mathcal{O}$ serves as a 3D patch, denoted as $\mathbf{x}_o^{(i)} \in P$. These serialized features are partitioned into non-overlapping local windows of a fixed size $K$. Controlled by a dilation rate $D$, the patch features at depth $o$ are padded and reconfigured into attention tokens:
\begin{equation}
\tilde{\mathbf{X}}_o = \operatorname{Flatten}(\operatorname{Transpose}(\operatorname{Reshape}(\hat{\mathbf{X}}_o))),
\end{equation}
where $\tilde{\mathbf{X}}_o \in \mathbb{R}^{B \times K \times C}$ represents the grouped window tokens, with $B$ and $C$ denoting the window count and channel dimension, respectively. A standard self-attention mechanism runs across all windows:
\begin{equation}
\mathbf{F}_o = \operatorname{Softmax}\left(\frac{\mathbf{Q} \mathbf{K}^\top}{\sqrt{d}} + \mathbf{B}\right) \mathbf{V}, \quad \text{with } \mathbf{Q},\mathbf{K},\mathbf{V} = \tilde{\mathbf{X}}_o \mathbf{W}_{q,k,v},
\end{equation}
where $\mathbf{W}_{q,k,v}$ are learnable projection weights, and $\mathbf{B}$ is the conditional positional encoding bias. The output tensor $\mathbf{F}_o \in \mathbb{R}^{B \times K \times C}$ gathers the updated patch representations across all $B$ windows at depth $o$. Finally, $\mathbf{F}_o$ is converted back to the linear array through $\operatorname{Reverse}(\cdot)$ and with padding elements removed through $\operatorname{Mask}(\cdot)$, before being down sampled to output the feature for each patch:
\begin{equation}
\mathbf{f}_o^{(i)} = h(\mathbf{x}_o^{(i)}) = \operatorname{Downsample}(\operatorname{Reverse}(\operatorname{Mask}(\mathbf{F}_o)))^{(i)}.
\end{equation}

\textbf{Feature aggregator.} Given the features $\mathbf{f}_o^{(i)}$ extracted by the backbone from all patches at depth $o$, we employ a cross-attention mechanism to adaptively pool them into a fixed set of tokens \citep{goswami2024salsa}. Let $\mathbf{K}_o = \mathbf{V}_o = \{\mathbf{f}_o^{(i)}\}_{i=1}^N \in \mathbb{R}^{N \times C}$ stack the patch features as the keys and values. Utilizing a learnable query matrix $\mathbf{Q}_\theta \in \mathbb{R}^{k \times d}$, where $k$ and $d$ denote the token count and channel dimension respectively, the adaptive pooling layer generates the initial tokens $\mathbf{T}_o \in \mathbb{R}^{k \times d}$:
\begin{equation}
\mathbf{T}_o = \text{Softmax}\left(\frac{\mathbf{Q}_\theta \mathbf{K}_o^\top}{\sqrt{d}}\right) \mathbf{V}_o.
\end{equation}
These pooled tokens are subsequently refined by a token fuser, which consists of a stack of four residual two-layer MLP blocks, yielding the intermediate tokens $\mathbf{H}_o \in \mathbb{R}^{k \times d}$.  $\mathbf{H}_o$ is then mapped through an MLP-Mixer \citep{tolstikhin2021mlp} performing consecutive token-mixing and channel-mixing:
\begin{equation}
\mathbf{U}_o = \mathbf{H}_o + \mathbf{W}_2 \sigma \left( \mathbf{W}_1 \text{LN}(\mathbf{H}_o)^\top \right)^\top,
\end{equation}
\begin{equation}
\mathbf{M}_o = \mathbf{U}_o + \mathbf{W}_4 \sigma \left( \mathbf{W}_3 \text{LN}(\mathbf{U}_o) \right),
\end{equation}
where $\text{LN}$ represents layer normalization, $\sigma$ is the non-linear activation function, and $\mathbf{W}_{1,2,3,4}$ denote the learnable weight parameters of the token-mixing and channel-mixing layers, respectively. Finally, the resulting matrix $\mathbf{M}_o \in \mathbb{R}^{k \times d}$ is flattened and $L_2$-normalized to yield the final global descriptor.

\textbf{Multi-scale patch-level self-supervised learning.} To mitigate the domain gap between aerial and ground point clouds, we extend the self-supervised 2D patch-level pretext task \citep{yun2022patch} to 3D point clouds. Based on the intuition that neighboring point cloud patches share similar semantics with the anchor patch, we design an asymmetric patch-level distillation framework with octree structures. Similar neighboring patches are selected and aggregated to provide multi-scale self-supervision for the central patch. This framework not only increases local structural consistency between point cloud patches, but also improves the distinctiveness of global features for retrieval.

Let $o \in \mathcal{O}$ denote the depth of an octree from a point cloud. For each non-empty central patch $\mathbf{x}_o^{(i)}$, we first retrieve its spatial neighboring patches $\mathcal{N}_o^{(i)}$. To select patches sharing similar semantics, we extract the raw feature $\mathbf{f}_{o}^{(i)} = h(\mathbf{x}_o^{(i)})$ through the backbone and compute the cosine similarity with its neighbors:
\begin{equation}
    s_o(i, j) = \frac{\mathbf{f}_{o}^{(i)\top} \mathbf{f}_{o}^{(j)}}{\|\mathbf{f}_{o}^{(i)}\|_2 \|\mathbf{f}_{o}^{(j)}\|_2}, \quad \forall j \in \mathcal{N}_o^{(i)}.
\end{equation}
Based on similarity scores, we filter and retain the most similar patches to construct a positive candidate group $\mathcal{P}_o^{(i)} \subset \mathcal{N}_o^{(i)}$. 

To mitigate the effects of potentially noisy patches in $\mathcal{P}_o^{(i)} $, the teacher network leverages cross-attention blocks to aggregate these neighboring features. Specifically, the anchor patch feature $\mathbf{f}_{\theta, o}^{(i)}$ serves as the query token ($\mathbf{q}$), while the anchor patch and its positive neighbors in $\mathcal{P}_o^{(i)}$ form the keys and values. Through this aggregation module, the network synthesizes the target representation $\mathbf{y}_{\tilde{\theta}, o}^{(i)}$. Our task aims to align the student feature $\mathbf{f}_{\theta, o}^{(i)}$ with the teacher representation $\mathbf{y}_{\tilde{\theta}, o}^{(i)}$.

\textbf{Training loss.} The self-supervised loss is formulated at multiple scales:

\begin{equation}
    \mathcal{L}_{\text{SSL}} = \sum_{o \in \mathcal{O}} \frac{1}{N_o} \sum_{i=1}^{N_o} \mathcal{D}\left( g_{\theta}\left(\mathbf{f}_{\theta, o}^{(i)}\right), \, \text{sg}\left( g_{\tilde{\theta}}\left(\mathbf{y}_{\tilde{\theta}, o}^{(i)}\right) \right) \right),
\end{equation}
where $N_o$ represents the total number of non-empty voxels at the octree depth $o$, $\mathcal{D}$ is the cross-entropy loss, and $\text{sg}(\cdot)$ is the stop-gradient operator. We utilize a self-distillation scheme to optimize the model. The student network parameterized by $\theta$ is optimized via gradient descent, whereas the teacher network parameters $\tilde{\theta}$ are updated via an Exponential Moving Average (EMA) of the student network, formulated as $\tilde{\theta} \leftarrow m\tilde{\theta} + (1-m)\theta$, where $m$ is the momentum coefficient. The projection head $g$ is implemented as an MLP followed by a Softmax to produce a probability distribution. $\mathcal{L}_{\text{SSL}}$ is designed to minimize the distance between a center point cloud patch and its similar neighboring patches.

The global triplet margin loss $\mathcal{L}_{\text{global}}$ learns an embedding space that minimizes the distance between the query point cloud feature $f(P_q)$ and its positive sample feature $f(P_p)$. 
\begin{equation}
\mathcal{L}_{\text{global}} = \max(\|f(P_q) - f(P_p)\|^2 - \|f(P_q) - f(P_n)\|^2 + m, 0).
\end{equation}
Here, $\|\cdot\|^2$ denotes the $L_2$ norm and $m$ is the margin. The total loss is composed of the two losses with a hyperparameter $\lambda$.

\begin{equation}
\mathcal{L}_{\text{total}} = \mathcal{L}_{\text{global}} + \lambda \mathcal{L}_{\text{SSL}}.
\end{equation}

\subsection{Re-ranking with expanded reciprocal neighbors}
\label{sec_3_3}

\begin{figure}[]
\centering
\includegraphics[width=0.95\linewidth, keepaspectratio]{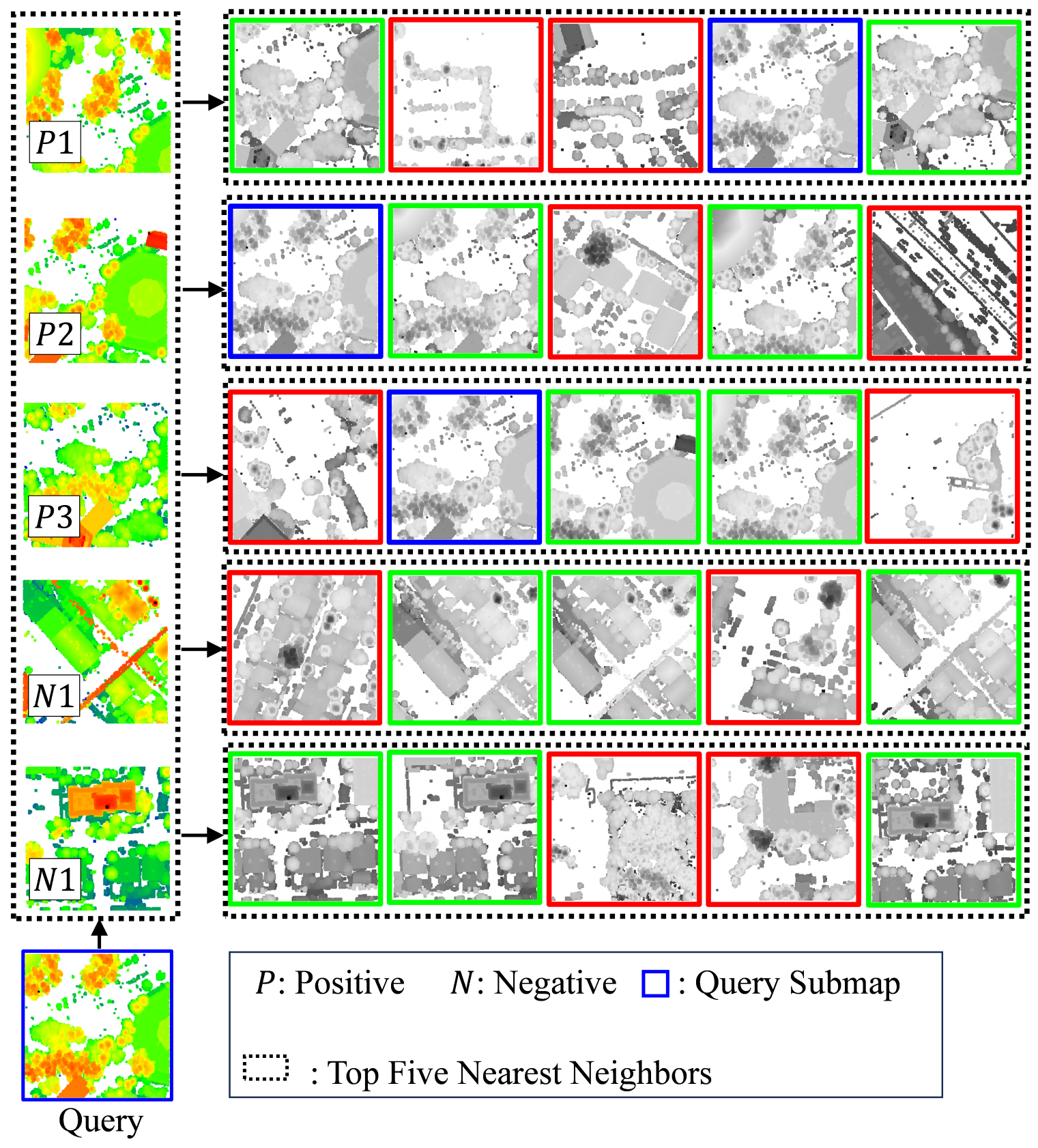}
\caption {Concept of reciprocal neighbors. Suppose a query whose top-5 neighbors consist of three positives ($P_1, P_2, P_3$) and two negatives ($N_1, N_2$). For each candidate, its own positive and negative samples are colorized in green and red, respectively. $P_1$, $P_2$, and $P_3$ are recognized as reciprocal neighbors because their own top-5 lists contain the query. In contrast, the negatives $N_1$ and $N_2$ are filtered out through this reciprocal mechanism.}
\label{fig_reciprocal}
\end{figure}

\begin{figure}[]
\centering
\includegraphics[width=0.9\linewidth, keepaspectratio]{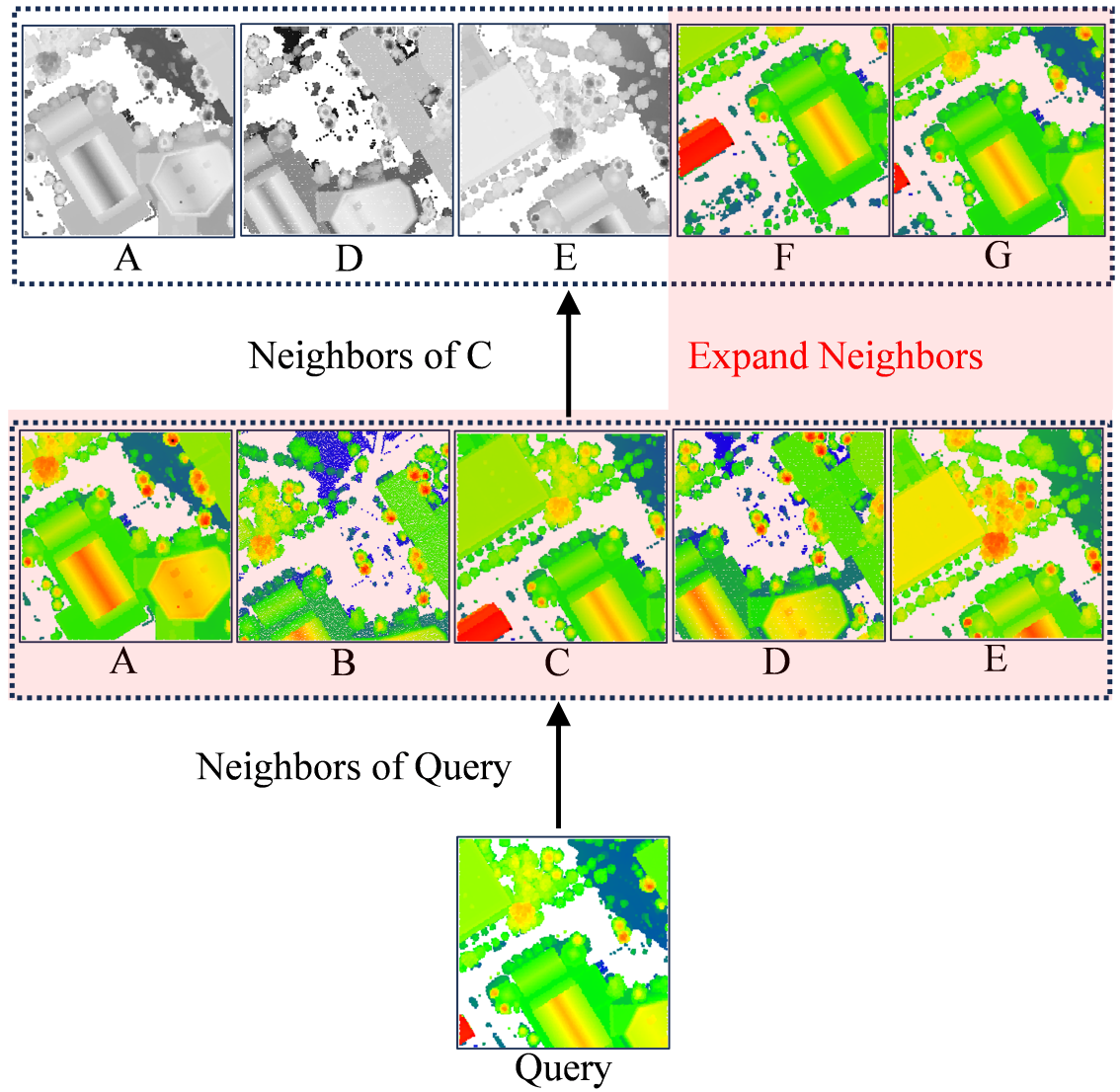}
\caption{Demonstration of neighborhood expansion. For a query point cloud $i$, its initial neighbor set contains $\{A, B, C, D, E\}$. Since neighbor $C$ retrieves its own neighbors $\{A, D, E, F, G\}$, $F$ and $G$ are also included through expansion.}
\label{fig_expansion}
\end{figure}

To mitigate the numerous false positives in the initial retrieval results, we introduce a training-free and lightweight re-ranking method. Given that the ALS database is continuously collected with uniform spatial overlaps, we utilize reciprocal neighbors \citep{zhong2017re} and perform neighborhood expansion to leverage this structured spatial distribution. In this section, we first formulate the reciprocal neighbors and then perform neighbor expansion. Cross distances are calculated using the refined features to update the final ranking. 

Figure ~\ref{fig_success} and Figure ~\ref{fig_failure} demonstrate the geographical distributions after re-ranking, where spatially close point clouds tend to cluster, regardless of whether they are correctly retrieved or not. Algorithm ~\ref{alg:rerank} summarizes our re-ranking pipeline, which requires no additional training and directly utilizes and refines global features.

\textbf{Reciprocal neighbors.} Let $\mathcal{S} = \mathcal{Q} \cup \mathcal{D}$ be the union of the ground query set $\mathcal{Q}$ and the aerial database $\mathcal{D}$. We compute the initial distances $d(i, j)$ between any two point cloud features $i, j \in \mathcal{S}$. For each point cloud $i$, its $k$-nearest neighbors are denoted as $\mathcal{N}(i, k)$. 

\begin{align}
d(i, j) = 1 - \frac{\langle f(i), f(j) \rangle}{\|f(i)\|_2 \|f(j)\|_2}, \quad \forall i, j \in \mathcal{S}.
\end{align}

\begin{align}
N(i, k) = \{j_1, j_2, \dots, j_k\}, |N(i, k)| = k. 
\end{align}

The reciprocal neighbors of $i$, denoted as $\mathcal{R}(i)$, require that any neighbor $j$ and $i$ are mutually ranked within each other's top-$k$ lists. Figure~\ref{fig_reciprocal} illustrates the concept of reciprocal neighbors.

\begin{align}
R(i) &= \{j \mid (j \in N(i, k)) \wedge (i \in N(j, k))\}.
\end{align}

\textbf{Neighbor expansion.} Since truly similar point clouds may fall outside $N(i, k)$ and thus be excluded from $R(i)$, we extend reciprocal neighbors to include the $k$-reciprocal neighbors of each element in $R(i)$, which is defined as the extended reciprocal neighbor set $E(i)$. The neighborhood expansion process is demonstrated in Figure ~\ref{fig_expansion}.

\begin{align}
E(i) &= R(i) \cup \bigcup_{j \in R(i)} R(j).
\end{align}

Based on the expanded neighborhood $\mathcal{E}(i)$, for each point cloud $i \in \mathcal{S}$ we refine the feature $f(i)$ to $f'(i)$ by averaging all elements within $E(i)$.

\begin{align}
f'(i) &= \begin{cases} 
\frac{1}{|E(i)|} \sum_{x \in E(i)} f(x), & \text{if }|E(i)| > 0, \\ 
f(i), & \text{if } |E(i)| = 0. 
\end{cases}
\end{align}

\textbf{Re-ranking.} Given a ground query $P_q \in \mathcal{Q}$ and an aerial database submap $M_i \in \mathcal{D}$, the final distance $d_{\text{final}}(P_q, M_i)$ is calculated using the refined features.

\begin{align}
d'(i, j) = 1 - \frac{\langle f'(i), f'(j) \rangle}{\|f'(i)\|_2 \|f'(j)\|_2}, \quad \forall i, j \in \mathcal{S},
\end{align}

\begin{equation}
\begin{aligned}
d_{final}(P_q, M_i) &= \frac{1}{|E(P_q)| + |E(M_i)|} \left[ \sum_{x \in E(P_q)} d'(x, M_i) \right. \\
                    & \left. + \sum_{y \in E(M_i)} d'(P_q, y) \right],
\end{aligned}
\end{equation}
where $d'(i, j)$ denotes the distance between the refined features of point clouds $i$ and $j$. The re-ranking is performed based on the final distance matrix $d_{final}$.

\begin{algorithm}[htbp]
\caption{Expanded Reciprocal Re-ranking}
\label{alg:rerank}
\begin{algorithmic}[1]
\renewcommand{\algorithmicrequire}{\textbf{Input:}}
\renewcommand{\algorithmicensure}{\textbf{Output:}}

\REQUIRE Features of query set $\mathcal{Q}$ and database $\mathcal{D}$ extracted by $f(\cdot)$; neighborhood parameter $k$
\ENSURE Final re-ranked distance matrix $d_{final}$

\STATE Let $\mathcal{S} = \mathcal{Q} \cup \mathcal{D}$. Compute $d(i, j)$ for all $i, j \in \mathcal{S}$.
\STATE Find $k$-nearest neighbors $N(i,k)$.
\STATE Find reciprocal neighbors $R(i)$.
\STATE Find expanded reciprocal neighbors $E(i)$.
\STATE Refine features $f'(i)$ 
\STATE Compute $d'(i, j)$ based on refined features $f'$.
\STATE \textbf{for} each ground query $P_q \in \mathcal{Q}$ and aerial database submap $M_i \in \mathcal{D}$ \textbf{do}
\STATE \quad Let $d_1 = \sum_{x \in E(P_q)} d'(x, M_i)$
\STATE \quad Let $d_2 = \sum_{y \in E(M_i)} d'(P_q, y)$
\STATE \quad $d_{final}(P_q, M_i) = \frac{d_1 + d_2}{|E(P_q)| + |E(M_i)|}$
\STATE \textbf{end for}
\end{algorithmic}
\end{algorithm}

\section{CS-Urban-Scenes dataset}
\label{sec_4}

\begin{figure*}
\centering
\includegraphics[width=0.9\textwidth, keepaspectratio]{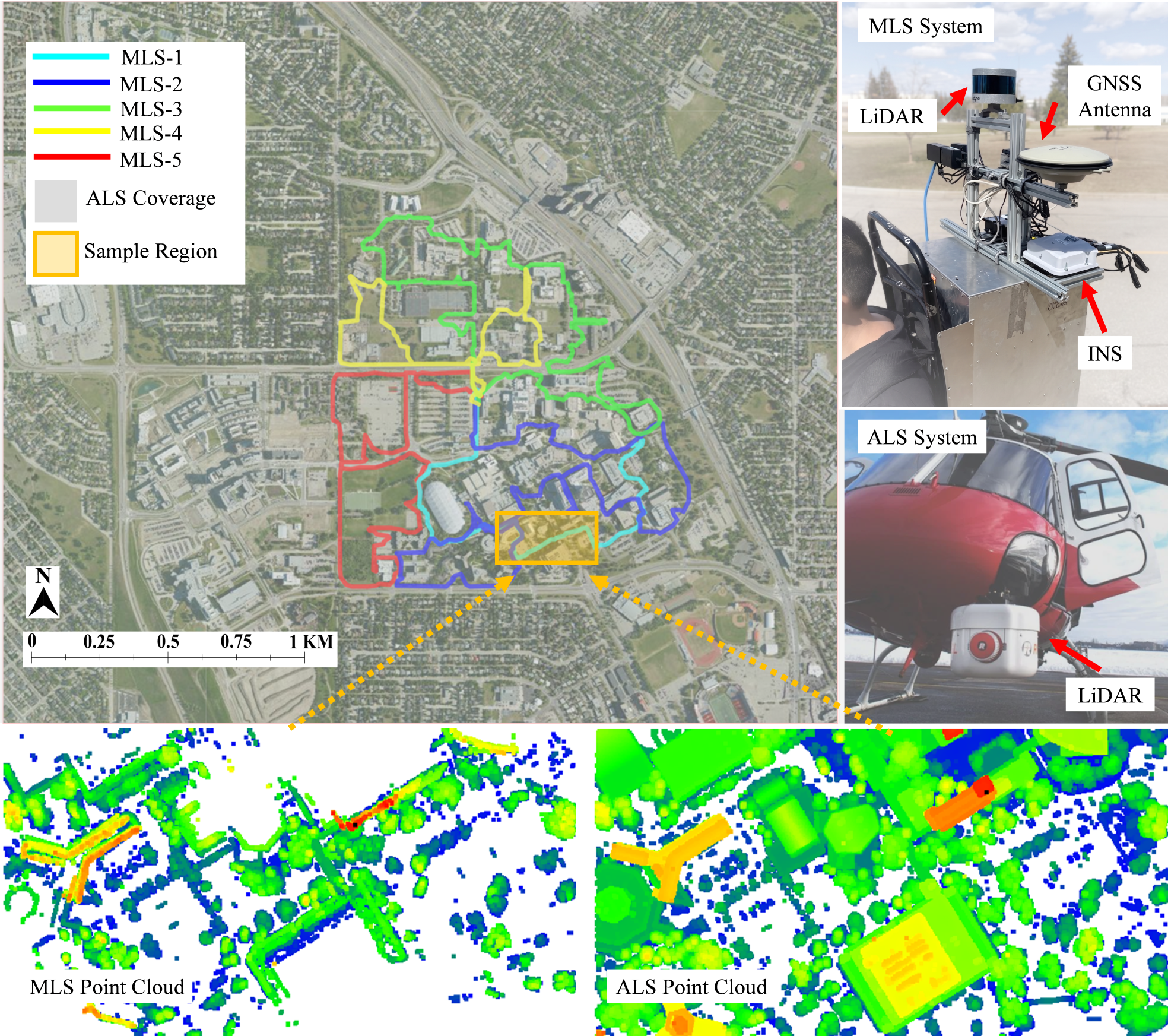}
\caption {Illustration of our CS-Urban-Scenes dataset. The coverage of the satellite imagery is consistent with that of the aerial point cloud database, spanning an area of 3 km by 2.4 km.}
\label{fig_traj}
\end{figure*}

Our multi-source heterogeneous dataset, CS-Urban-Scenes, integrates Mobile Laser Scanning (MLS) and Aerial Laser Scanning (ALS) data from urban areas in Calgary. Figure ~\ref{fig_traj} presents an overview of the CS-Urban-Scenes dataset, where MLS trajectories are mapped onto satellite imagery sharing same geographic coverage with the aerial point cloud database.\footnote{Image of the ALS system: \url{https://www.riegl.com/en-canada/products/detail/riegl-vp-1}} Sensor configurations of the backpacking mobile mapping system are also provided. This dataset is constructed to benchmark aerial-ground LiDAR place recognition algorithms under large-scale urban scenarios. It captures a large urban area of 7.2~$\text{km}^2$ with 18.1~\text{km} of trajectories, where the ground-truth poses are computed via Post-Processing Kinematic (PPK).

\textbf{Study area.} The data were collected around the main campus of the University of Calgary. The dataset captures a structured urban environment, mainly containing buildings, trees, and street infrastructure. While the MLS point clouds are limited within 1.5 $km^2$, the ALS point clouds cover an area of 7.2 $km^2$ as illustrated in Figure ~\ref{fig_traj}.

\textbf{Ground data acquisition.} The MLS point clouds were acquired using a backpack mobile mapping system, integrated with a Velodyne VLP-16 LiDAR, a NovAtel PwrPak7-E1 GNSS/INS system, and a NovAtel GPS-702-GG-HV antenna. The LiDAR provides an effective range of 100 m and a point accuracy of $\pm 3$~cm. When there is no GNSS outage, the GNSS/INS solution achieves a positional accuracy of 1 cm and an attitude accuracy of 0.01 degree after Inertial Explorer post-processing, yielding MLS point cloud maps with an accuracy of 5 cm. Five subsets are collected with a total trajectory length of 18.1 km. 

\textbf{Aerial data acquisition.} Acquired in 2020, the Calgary ALS point clouds are provided by the Government of Canada \footnote{Natural Resources Canada: \url{https://open.canada.ca/data/en/dataset/7069387e-9986-4297-9f55-0288e9676947}}. They have an average point density of 14 pts/$m^2$, with horizontal and vertical accuracies of 0.2 m and 0.1 m, respectively. We selected the ALS point cloud data covering the university area, which spans 7.2 $km^2$ (3.0 km $\times$ 2.4 km).

\textbf{Dataset preparation.} Ground submaps of radius 100 m were extracted from the MLS point clouds every 2 meters along the trajectory, yielding a total of 9,068 submaps. Ground points for each submap were removed using CSF \citep{zhang2016easy}; the remaining non-ground points were then downsampled to 4,096 and normalized to the range [-1, 1]. Aerial point clouds were partitioned into $100 \times 100\text{ m}^2$ submaps with a stride of 19 m on both east and north directions. These aerial submaps were also ground-removed, downsampled and normalized. The total number of aerial submaps is 35,212.

\textbf{Train-test split.} To maintain an approximate train-to-test ratio of 4:1, we assigned MLS-2, MLS-3, MLS-5, and the latter part of MLS-4 to the training set, while setting MLS-1 and the first part of MLS-4 as the test set. This resulted in 7,242 submaps for training and 1,826 for testing. We use all the aerial submaps during training. We define a successful retrieval when the distance between the ground and aerial submaps is less than 100 m.

\section{Experiments}
\label{sec_5}

Our method consists of a retrieval stage followed by a re-ranking stage that refines the initial retrieval results without further training. In this section, we first introduce the experimental settings and then evaluate the two stages from both quantitative and qualitative perspectives. Finally, we provide ablation studies and further analysis of the results.

\subsection{Settings}

\begin{table}[!t]
\centering
\caption{Aerial-ground LiDAR place recognition dataset.}
\label{tab_datasets}
\setlength{\tabcolsep}{4pt} 
\small %
\resizebox{\columnwidth}{!}{
\begin{tabular}{lcc}
\toprule
Dataset & CS-Campus3D & CS-Urban-Scenes (Ours) \\ \midrule
Environment & Campus & Urban \\ 
Platform & Robot / Airplane & Backpack / Airplane \\ 
Length / Coverage & 7.8 km / 5.5 km$^2$ & 18.1 km / 7.2 km$^2$ \\ 
GT Trajectories & U-blox GPS & NovAtel PPK \\ 
Ground Submaps & \makecell[c]{6,167 / 1,538\\(train / test)} & \makecell[c]{7,242 / 1,826\\(train / test)} \\ 
Aerial Submaps & 27,520 (Database) & 35,212 (Database) \\ 
Submap Size & 100 m & 100 m \\ 
Positive Threshold & 100 m & 100 m \\ \bottomrule
\end{tabular}
}
\end{table}

\textbf{Retrieval baselines.}
For the evaluation of initial retrieval, we train the MinkLoc3D \citep{komorowski2021minkloc3d}, CrossLoc3D \citep{guan2023crossloc3d}, and HOTFormerLoc \citep{griffiths2025hotformerloc} models on both the CS-Campus3D and CS-Urban-Scenes datasets. MinkLoc3D employs sparse voxelization and convolutions to extract local features from point clouds. CrossLoc3D utilizes an iterative refinement process to obtain local features. HOTFormerLoc introduces relay tokens to facilitate multi-scale global-local interactions.

\textbf{Re-ranking baselines.}
For the re-ranking stage, we extract global features using CrossLoc3D, HOTFormerLoc, and our proposed model, and re-rank the retrieval results using various re-ranking methods, including $k$-reciprocal (KR) \citep{zhong2017re}, ECN \citep{sarfraz2018pose}, Cheb-GR \citep{yang2025cheb}, $\alpha$-QE \citep{radenovic2018fine}, and SuperGlobal (SG) \citep{shao2023global}. $k$-reciprocal (KR) re-ranking refines retrieval scores by combining the original distance with a Jaccard distance, while the ECN refines initial feature distances using the contextual similarity of the top nearest neighbors. $\alpha$-QE expands each query with a distance-weighted average of its top neighbors, Cheb-GR refines query and database descriptors via graph-based propagation with adaptive thresholding, and SuperGlobal applies top descriptor augmentation followed by maximum-descriptor aggregation reranking. 

\textbf{Experimental datasets.} We validate our retrieval and re-reranking algorithms on the CS-Campus3D \citep{guan2023crossloc3d} and our self-collected CS-Urban-Scenes datasets. While CS-Campus3D relies on a GPS module and uses individual LiDAR frames as queries, CS-Urban-Scenes utilizes NovAtel PPK processing and merges frames into submaps. Section~\ref{sec_4} provides more details on the CS-Urban-Scenes, and Table ~\ref{tab_datasets} summarizes the comparison between two datasets.

\textbf{Evaluation metrics.} In the retrieval stage, we employ $\textit{Average Recall}@N$ ($\textit{AR}@N$) and $\textit{AR}@1\%$, where $\textit{AR}@N$ measures the percentage of positives within the top $N$ predictions (or top $1\%$ of the database). For re-ranking, we additionally report the mean Average Precision (mAP).

\textbf{Training parameters.} Our training is implemented in PyTorch and is conducted on an NVIDIA H100 GPU provided by the Digital Research Alliance of Canada \footnote{Digital Research Alliance of Canada: \url{https://www.alliancecan.ca/}}. The network warms up for the first 20 epochs, after which the teacher network begins updating via EMA. The training spans 300 epochs with a learning rate of $5\times10^{-4}$ and a loss weighting hyperparameter $\lambda = 0.1$. For the self-supervised parameters, the teacher temperature, student temperature, and EMA momentum are configured as $0.04$, $0.1$, and $0.996$, respectively. 

\subsection{Quantitative results}
Quantitatively, we conduct separate evaluations to initial retrieval and re-ranking results. For comparison of retrieval performance with SOTA algorithms, average recall rates ($AR@N$) are utilized. To evaluate the re-ranking stage, Mean Average Precision (mAP) is also adopted as the evaluation metric.

\begin{figure}[]
\centering
\includegraphics[width=0.8\linewidth, keepaspectratio]{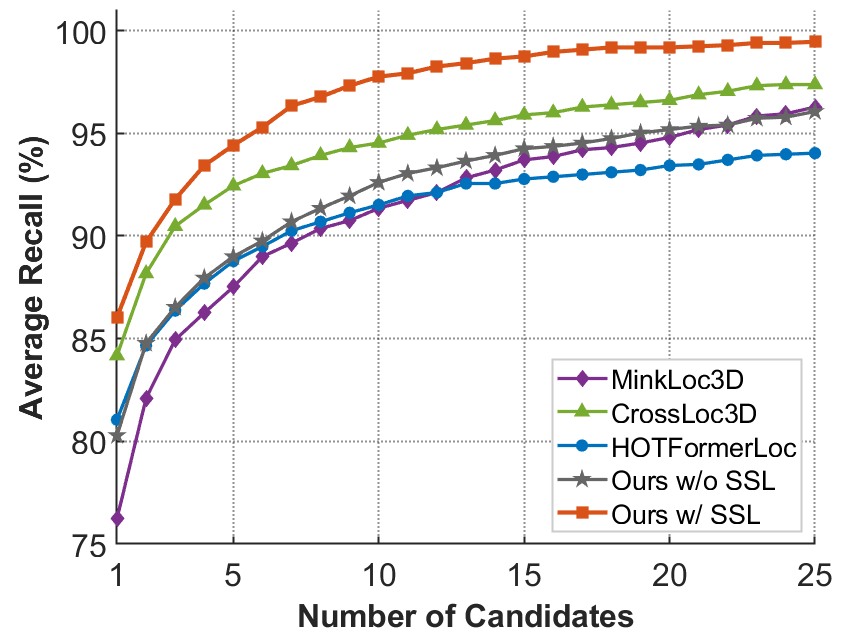}
\caption {Retrieval recall curves of SOTA algorithms on CS-Urban-Scenes datasets.}
\label{fig_curve}
\end{figure}

\begin{table}[!t]
\centering
\caption{Retrieval results on CS-Campus3D.}
\label{tab_results_campus3d}
\small
\setlength{\tabcolsep}{4pt} 
\begin{tabular}{lcc} 
\toprule
Method & AR@1  & AR@1\%  \\
\midrule
PointNetVLAD & 19.1 & 43.6 \\
MinkLoc3D    & 55.4 & 85.2 \\
CrossLoc3D   & 72.3 & 86.1 \\
HOTFormerLoc & 78.4 & 89.5 \\
\textbf{Ours w/o SSL} & 75.5 & 90.8 \\
\textbf{Ours w/ SSL} & \textbf{81.1} & \textbf{91.3} \\
\bottomrule
\end{tabular}
\end{table}

Table~\ref{tab_results_campus3d} reports the retrieval results on the CS-Campus3D dataset. As evaluated, our model achieves the best performance across all metrics with 81.1\% in $AR@1$ and 91.3\% in $AR@1\%$. Furthermore, comparing our model with and without SSL highlights the effectiveness of patch-level self-supervised learning, which yields a 5.6\% improvement in $AR@1$. Figure~\ref{fig_curve} presents the top-25 average recall values of SOTA methods on the CS-Urban-Scenes dataset. Our method outperforms all other algorithms and the inclusion of patch-level self-supervised learning increases AR@1 by 5.7\%. Based on the prior knowledge that neighboring patches share similar semantics with the anchor patch, we design a multi-scale patch-level self-supervised learning scheme, which enhances global feature representations to increase retrieval accuracy.

\begin{table*}[!t]
\centering
\caption{Improvement in rank-$N$ and mAP with our Expanded Reciprocal (ER) re-ranker on different retrieval methods and datasets. "Initial" and "ER" denote the retrieval results before and after re-ranking, respectively, with the best results in bold. We report the results with the neighborhood parameter of $k=10$ for CS-Campus3D and $k=30$ for CS-Urban-Scenes.}
\label{tab:rr_results}
{\small
\begin{tabular}{l|cccccc|cccccc}
\hline
 & \multicolumn{6}{c|}{CS-Campus3D} & \multicolumn{6}{c}{CS-Urban-Scenes} \\
 & \multicolumn{2}{c}{AR@1} & \multicolumn{2}{c}{AR@5} & \multicolumn{2}{c|}{mAP} & \multicolumn{2}{c}{AR@1} & \multicolumn{2}{c}{AR@5} & \multicolumn{2}{c}{mAP} \\
Retrieval & Initial & ER & Initial & ER & Initial & ER & Initial & ER & Initial & ER & Initial & ER \\ \hline
CrossLoc3D   & 72.3 & \textbf{78.2} & 79.4 & \textbf{80.5} & 53.0 & \textbf{64.6} & 84.2 & \textbf{94.5} & 92.4 & \textbf{97.3} & 26.4 & \textbf{49.1} \\
HOTFormerLoc & 78.4 & \textbf{83.4} & 84.3 & \textbf{86.1} & 56.9 & \textbf{68.3} & 81.1 & \textbf{85.4} & 88.8 & \textbf{90.0} & 30.6 & \textbf{50.8} \\
\textbf{Ours}         & 81.1 & \textbf{86.0} & 85.5 & \textbf{87.9} & 63.6 & \textbf{71.9} & 86.0 & \textbf{96.2} & 94.4 & \textbf{98.8} & 33.4 & \textbf{57.9} \\ \hline
\end{tabular}
}
\end{table*}

\begin{table*}[!t]
\centering
\caption{Comparison on CS-Campus3D dataset with different re-ranking methods. Best results are in bold.}
\label{tab:reranking_comparison_campus}
{\small
\setlength{\tabcolsep}{3.2pt}
\begin{tabular}{l|cccccc|cccccc|cccccc}
\hline
 & \multicolumn{6}{c|}{AR@1} & \multicolumn{6}{c|}{AR@5} & \multicolumn{6}{c}{mAP} \\
Retrieval & KR & ECN & Cheb & $\alpha$-QE & SG & ER & KR & ECN & Cheb & $\alpha$-QE & SG & ER & KR & ECN & Cheb & $\alpha$-QE & SG & ER \\ \hline
CrossLoc3D   & 71.0 & 73.1 & 71.7 & 73.9 & 71.0 & \textbf{78.2} & 76.0 & 77.3 & 77.2 & 76.1 & 75.5 & \textbf{80.5} & 55.2 & 56.0 & 53.9 & 61.3 & 55.6 & \textbf{64.6} \\
HOTFormerLoc & 77.9 & 80.6 & 80.6 & 80.0 & 79.0 & \textbf{83.4} & 82.4 & \textbf{86.2} & 85.1 & 82.9 & 82.5 & 86.1 & 57.6 & 59.5 & 63.2 & 63.9 & 63.8 & \textbf{68.3} \\
\textbf{Ours}         & 80.8 & 83.2 & 82.7 & 81.9 & 81.4 & \textbf{86.0} & 84.1 & 87.4 & 85.8 & 84.0 & 84.1 & \textbf{87.9} & 64.5 & 66.2 & 68.9 & 67.9 & 68.3 & \textbf{71.9} \\ \hline
\end{tabular}
}

\vspace{10pt} %

\caption{Comparison on CS-Urban-Scenes dataset with different re-ranking methods. Best results are in bold.}
\label{tab:reranking_comparison_urban}
{\small
\setlength{\tabcolsep}{3.2pt}
\begin{tabular}{l|cccccc|cccccc|cccccc}
\hline
 & \multicolumn{6}{c|}{AR@1} & \multicolumn{6}{c|}{AR@5} & \multicolumn{6}{c}{mAP} \\
Retrieval & KR & ECN & Cheb & $\alpha$-QE & SG & ER & KR & ECN & Cheb & $\alpha$-QE & SG & ER & KR & ECN & Cheb & $\alpha$-QE & SG & ER \\ \hline
CrossLoc3D   & 91.2 & 85.9 & 84.6 & 77.8 & 75.8 & \textbf{94.5} & 94.6 & 92.2 & 92.1 & 86.1 & 83.0 & \textbf{97.3} & 34.9 & 27.9 & 27.1 & 34.7 & 27.8 & \textbf{49.1} \\
HOTFormerLoc & 80.5 & 80.7 & 84.7 & 81.0 & 80.1 & \textbf{85.4} & 88.6 & 89.7 & 89.9 & 84.6 & 84.5 & \textbf{90.0} & 31.9 & 45.1 & 34.4 & 43.1 & 43.0 & \textbf{50.8} \\
\textbf{Ours}         & 88.3 & 85.8 & 90.4 & 83.8 & 85.2 & \textbf{96.2} & 95.7 & 96.6 & 96.6 & 87.5 & 88.7 & \textbf{98.8} & 35.2 & 48.2 & 37.4 & 46.7 & 46.1 & \textbf{57.9} \\ \hline
\end{tabular}
}
\end{table*}

To verify the efficacy of our proposed Expanded Reciprocal (ER) algorithm, we conduct comprehensive evaluations across different datasets and retrieval networks. Table~\ref{tab:rr_results} presents the performance improvement before and after applying our ER re-ranker. It is observed that the utilization of ER yields substantial improvements across all metrics for all retrieval methods. On the CS-Urban-Scenes dataset, our ER algorithm increases the mAP of CrossLoc3D from 26.4\% to 49.1\% and mAP of our model from 33.4\% to 57.9\%, nearly doubling the initial performance.

Table~\ref{tab:reranking_comparison_campus} and Table~\ref{tab:reranking_comparison_urban} compare our ER algorithm against five re-ranking methods across different retrieval networks and datasets. On the CS-Campus3D dataset (Table~\ref{tab:reranking_comparison_campus}), most re-ranking methods bring improvements over the initial results. For example, when utilizing global features from our model, other methods achieve an mAP between 64.5\% and 68.9\%, while our ER algorithm achieves 71.9\%. On the CS-Urban-Scenes dataset (Table~\ref{tab:reranking_comparison_urban}), other methods show limitations. For CrossLoc3D features, re-rankers like $\alpha$-QE, and SuperGlobal fail to maintain the initial AR@1 accuracy of 84.2\%. In contrast, our ER algorithm increases the AR@1 to 94.5\%. Furthermore, based on our retrieval features, the highest mAP achieved by other methods is 48.2\% from ECN, whereas our ER algorithm reaches 57.9\%, outperforming it by of 9.7\%. By exploiting the structured spatial distribution of ALS point clouds, the expansion of reciprocal neighbors refines global features using neighborhood information, which mitigates false positives and significantly boosts the AR@1 accuracy without extra network training.

\subsection{Qualitative results}
To visually contrast the initial retrieval with the re-ranking, we structure our qualitative results into two parts. First, we map the retrieval recalls back onto the trajectories and colorize them according to their recalls ($R@1, 5, 10, 15, 20, 25$ and $>25$). Second, we select representative scenarios to demonstrate successful re-ranking. We visualize the positions of the top-25 candidates on aerial images and the top-5 point clouds before and after the re-ranking.

\begin{figure*}
\centering
\includegraphics[width=0.95\textwidth, keepaspectratio]{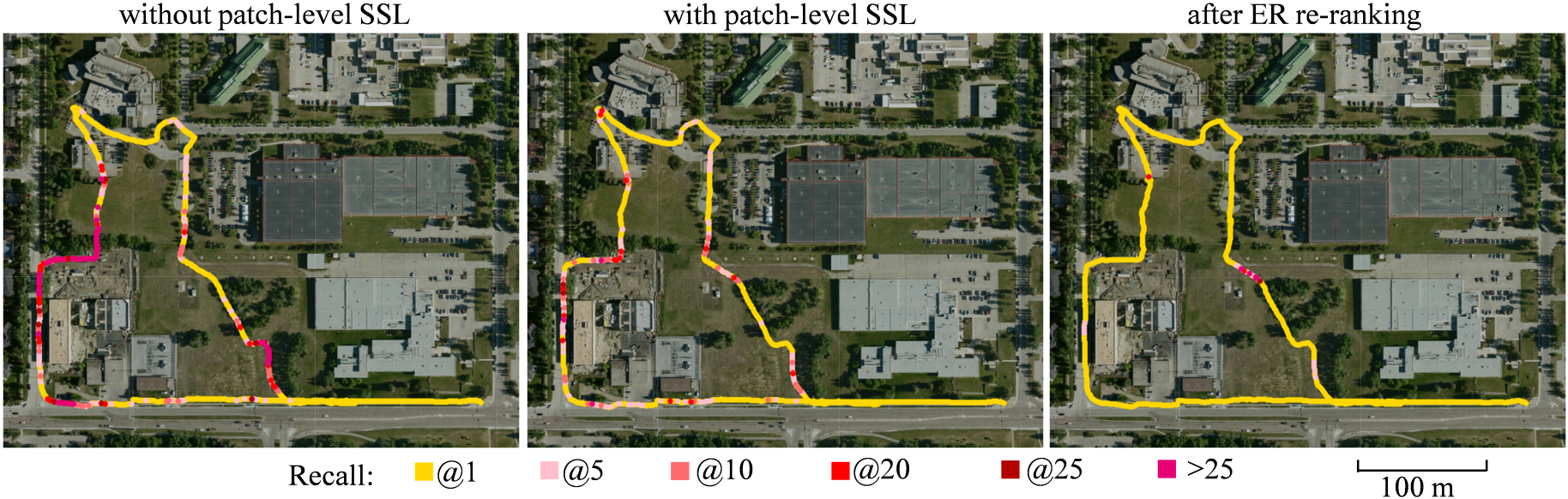}
\caption {Recall maps along the trajectories of the test set part of MLS-4 dataset. Trajectory points are colorized based on their categories (from $R@1$ to $R > 25$). The expanded reciprocal (ER) re-ranking is conducted based on the global features from the model with patch-level self-supervised learning (SSL).}
\label{fig_recallmap1}
\end{figure*}

\begin{figure*}
\centering
\includegraphics[width=0.95\textwidth, keepaspectratio]{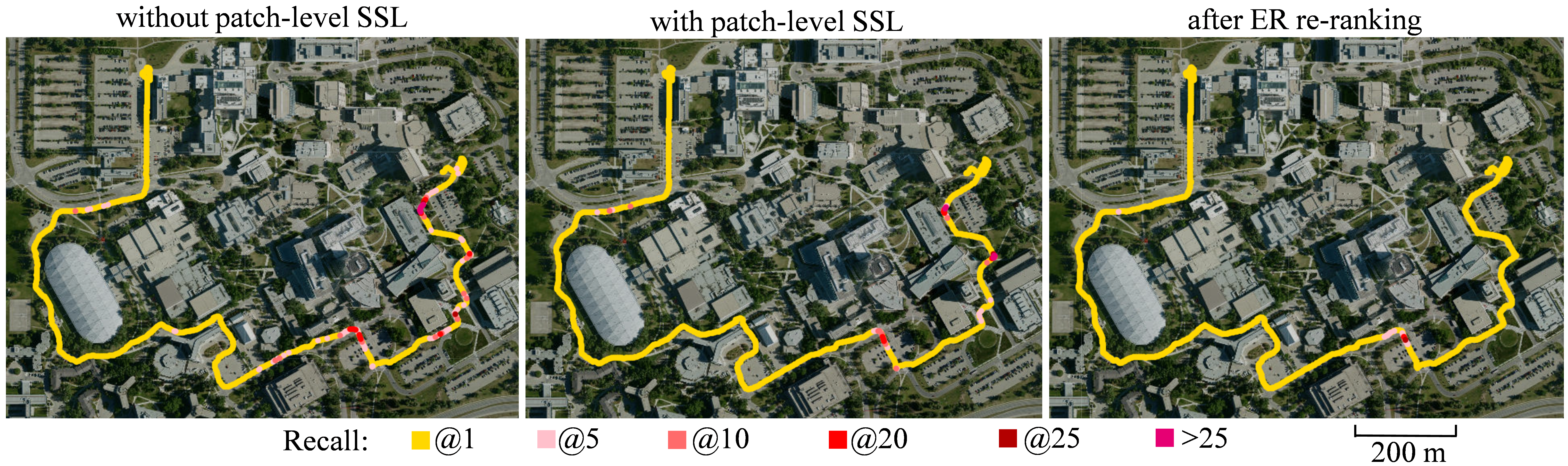}
\caption {Recall maps along the trajectories of the MLS-1 dataset. Trajectory points are colorized based on their categories (from $R@1$ to $R > 25$). The expanded reciprocal (ER) re-ranking is conducted based on the global features from the model with patch-level self-supervised learning (SSL).}
\label{fig_recallmap2}
\end{figure*}

Figures ~\ref{fig_recallmap1} and ~\ref{fig_recallmap2} illustrate the recall results for the MLS-4-1 and MLS-1 datasets, respectively. We contrast the results under three configurations: the baseline model without patch-level self-supervised learning (SSL), the model with patch-level SSL, and the results after Expanded Reciprocal (ER) re-ranking. The ER re-ranking is applied on the global features from the model with patch-level SSL. Compared to baseline, integrating patch-level SSL successfully increases retrieval recall, resulting in a higher number of $R@1$ points. Furthermore, when our ER re-ranking is utilized, the trajectory is almost dominated by $R@1$.

\begin{figure*}
\centering
\includegraphics[width=0.9\textwidth, keepaspectratio]{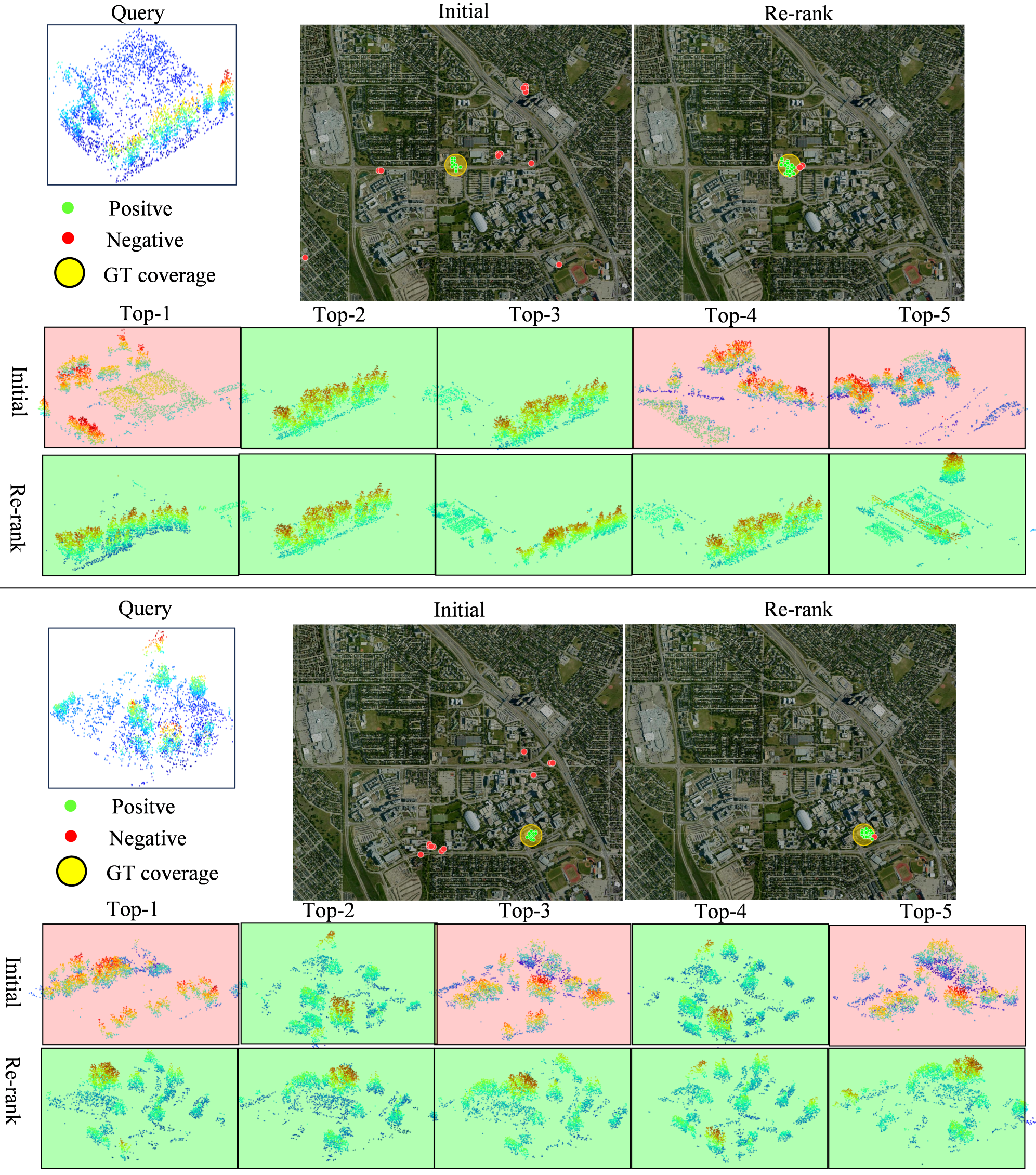}
\caption {Successful re-ranking queries. Green and red represent positive and negative candidates, respectively. After re-ranking, true positives are successfully promoted to rank-1 and the top-25 candidates cluster on the aerial image.}
\label{fig_success}
\end{figure*}

Figure~\ref{fig_success} demonstrates successful re-ranking scenarios using our proposed Expanded Reciprocal (ER) re-ranking algorithm. Two representative queries are selected for visualization. In aerial imagery, the 100-meter ground-truth radius is indicated by a yellow circle. The positions of the top-25 retrieved candidates before and after re-ranking are plotted, with green and red dots representing positive and negative candidates. Additionally, we visualize the top-5 retrieved point clouds in both configurations, rendered in green (positive) and red (negative). By expanding reciprocal neighbors, true positives are promoted to rank-1, and the precision of top-5 is enhanced. Although the initial retrieved candidates scatter across the aerial database, they tend to cluster around the query center after re-ranking.

\subsection{Ablation studies}

\begin{table}[!t]
\centering
\caption{Ablation studies on the number of positive patches.}
\label{tab_patch_number}
\small
\setlength{\tabcolsep}{4pt}
\begin{tabular}{ccccc} 
\toprule
& \multicolumn{2}{c}{CS-Campus3D} & \multicolumn{2}{c}{CS-Urban-Scenes} \\
\# Pos. Patches & AR@1  & AR@1\%  & AR@1  & AR@1\%  \\
\midrule
$\times$ & 75.5 & 90.8 & 80.3 & 96.1 \\ 
\midrule
2 & 79.3 & \textbf{93.2} & 85.7 & 98.4 \\
4 & \textbf{81.1} & 91.3 & \textbf{86.0} & \textbf{99.5} \\ 
6 & 80.5 & 91.7 & 84.7 & 98.1 \\
8 & 77.8 & 90.9 & 81.2 & 97.7 \\
\bottomrule
\end{tabular}
\end{table}

\begin{table}[!t]
\centering
\caption{Ablation studies on the scale of patch-level learning.}
\label{tab_patch_scale}
\small
\setlength{\tabcolsep}{3pt}
\begin{tabular}{*{3}{w{c}{1.25em}}cccc}
\toprule
\multicolumn{3}{c}{Octree Depth} & \multicolumn{2}{c}{CS-Campus3D} & \multicolumn{2}{c}{CS-Urban-Scenes} \\
4 & 3 & 2 & AR@1  & AR@1\%  & AR@1  & AR@1\%  \\
\midrule
$\times$ & $\times$ & $\times$ & 75.5 & 90.8 & 80.3 & 96.1 \\
$\checkmark$ & $\times$ & $\times$ & 78.9 & 89.1 & 84.2 & 98.1 \\
$\times$ & $\checkmark$ & $\times$ & 79.1 & \textbf{92.3} & 82.3 & 94.9 \\ 
$\times$ & $\times$ & $\checkmark$ & 80.1 & 90.7 & 83.4 & 97.6 \\ 
\midrule
$\checkmark$ & $\checkmark$ & $\checkmark$ & \textbf{81.1} & 91.3 & \textbf{86.0} & \textbf{99.5} \\ 
\bottomrule
\end{tabular}
\end{table}

\begin{figure}[]
\centering
\includegraphics[width=\linewidth, keepaspectratio]{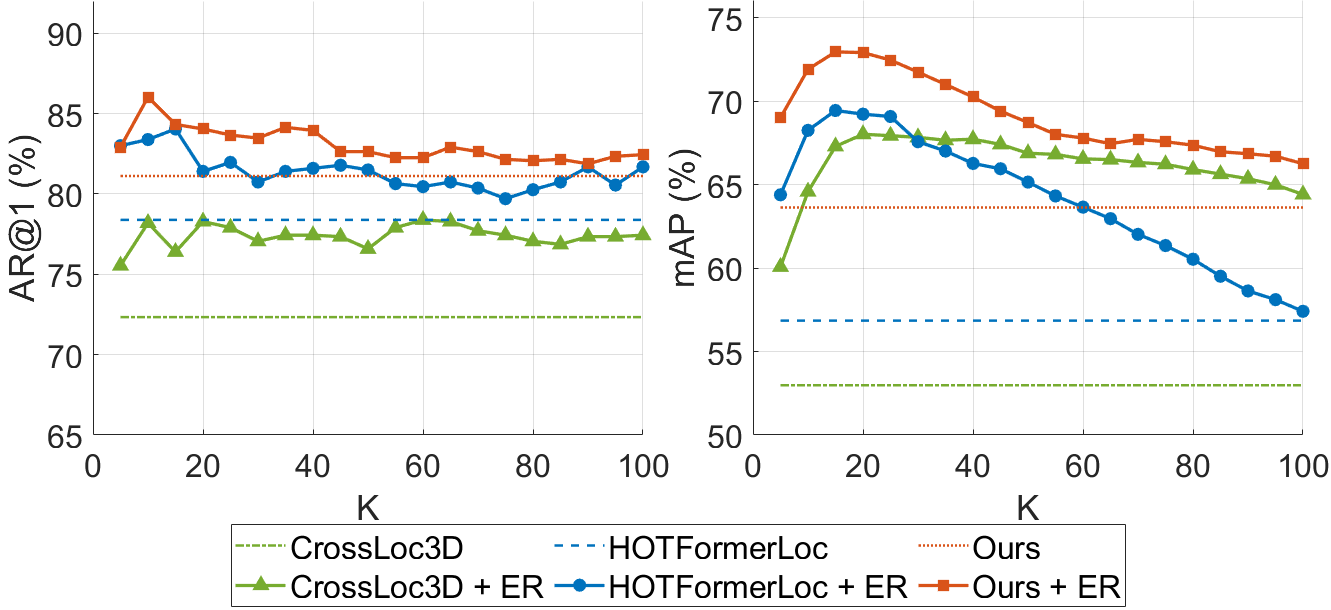}
\caption {Parameter sensitivity of the neighborhood size $k$ in the ER re-ranking method on CS-Campus3D.}
\label{fig_param1}
\end{figure}

\begin{figure}[]
\centering
\includegraphics[width=\linewidth, keepaspectratio]{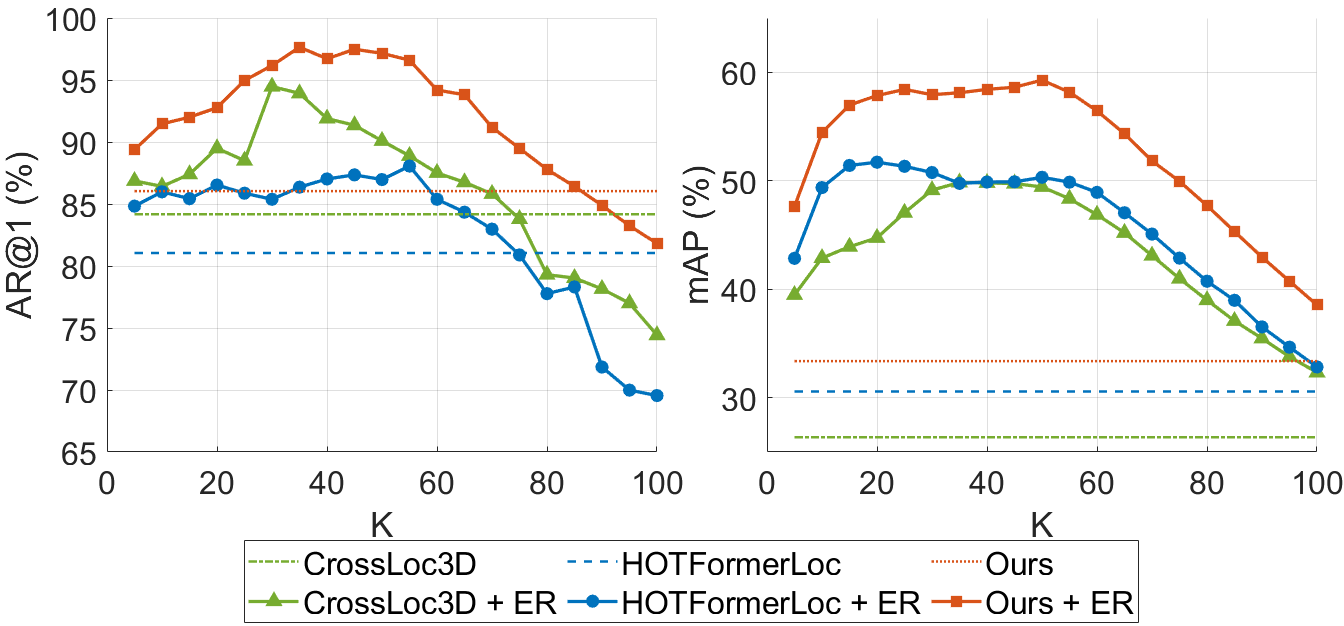}
\caption {Parameter sensitivity of the neighborhood size $k$ in the ER re-ranking method on CS-Urban-Scenes.}
\label{fig_param2}
\end{figure}

\textbf{Number of positive patches.} We evaluate the impact of the number of positive patches ($|\mathcal{P}_o^{(i)}|$) in the self-supervised module. As summarized in Table~\ref{tab_patch_number}, introducing positive patches consistently increased retrieval accuracy. Setting the number of positive patches to 4 achieves the best performance, with the AR@1 reaching 81.1\% on CS-Campus3D and 86.0\% on CS-Urban-Scenes. However, further increasing this number to 6 or 8 leads to a slight performance degradation, as a larger neighborhood within the octree introduces dissimilar patches or noise. 

\textbf{Scale of patch-level learning.} Table~\ref{tab_patch_scale} evaluates the impact of multi-scale patch-level learning. The results indicate that utilizing self-supervision at any octree depth improves the baseline. Besides, leveraging all three depths delivers the highest performance, which demonstrates that the multi-scale module captures both local and large-scale semantics.

\textbf{Number of neighbors in re-ranking.} Ablation studies are also conducted on the re-ranking stage . Figures~\ref{fig_param1} and~\ref{fig_param2} show how the neighborhood parameter $k$ affects performance. First, the results demonstrate the compatibility of our method. On both datasets, using any of the three baseline methods with the ER algorithm boosts performance. Second, the optimal choice for $k$ differs on the dataset. For the campus, all curves peak at $k=10$. In contrast, for the urban scene, the best AR@1 is achieved at $k=35$ and the highest mAP is reached at $k=50$.

\subsection{Analyses}

\begin{table}[!t]
\centering
\caption{Comparison of the efficiency of different retrieval methods.}
\label{tab:retrieval_time}
\small
\setlength{\tabcolsep}{4pt}
\begin{tabular}{ccc}
\toprule
Method & Parameters (M) & Inference Time (ms) \\
\midrule
MinkLoc3D    & 1.1  & 1.3 \\
CrossLoc3D   & 15.4 & 2.2 \\
HOTFormerLoc & 35.4 & 5.9 \\
Ours         & 34.5 & 4.7 \\
\bottomrule
\end{tabular}
\end{table}

\begin{table}[!t]
\centering
\caption{Comparison of the efficiency of different re-ranking methods.}
\label{tab:reranking_time}
\small
\setlength{\tabcolsep}{4pt}
\begin{tabular}{cc}
\toprule
Method & Processing Time (s) \\
\midrule
$k$-reciprocal & 26.42 \\
ECN            & 21.57 \\
$\alpha$-QE    & 17.17 \\
Cheb-GR        & 3.33  \\
SuperGlobal    & 2.60  \\
ER(Ours)            & 12.33 \\
\bottomrule
\end{tabular}
\end{table}

\begin{figure}[]
\centering
\includegraphics[width=0.9\linewidth, keepaspectratio]{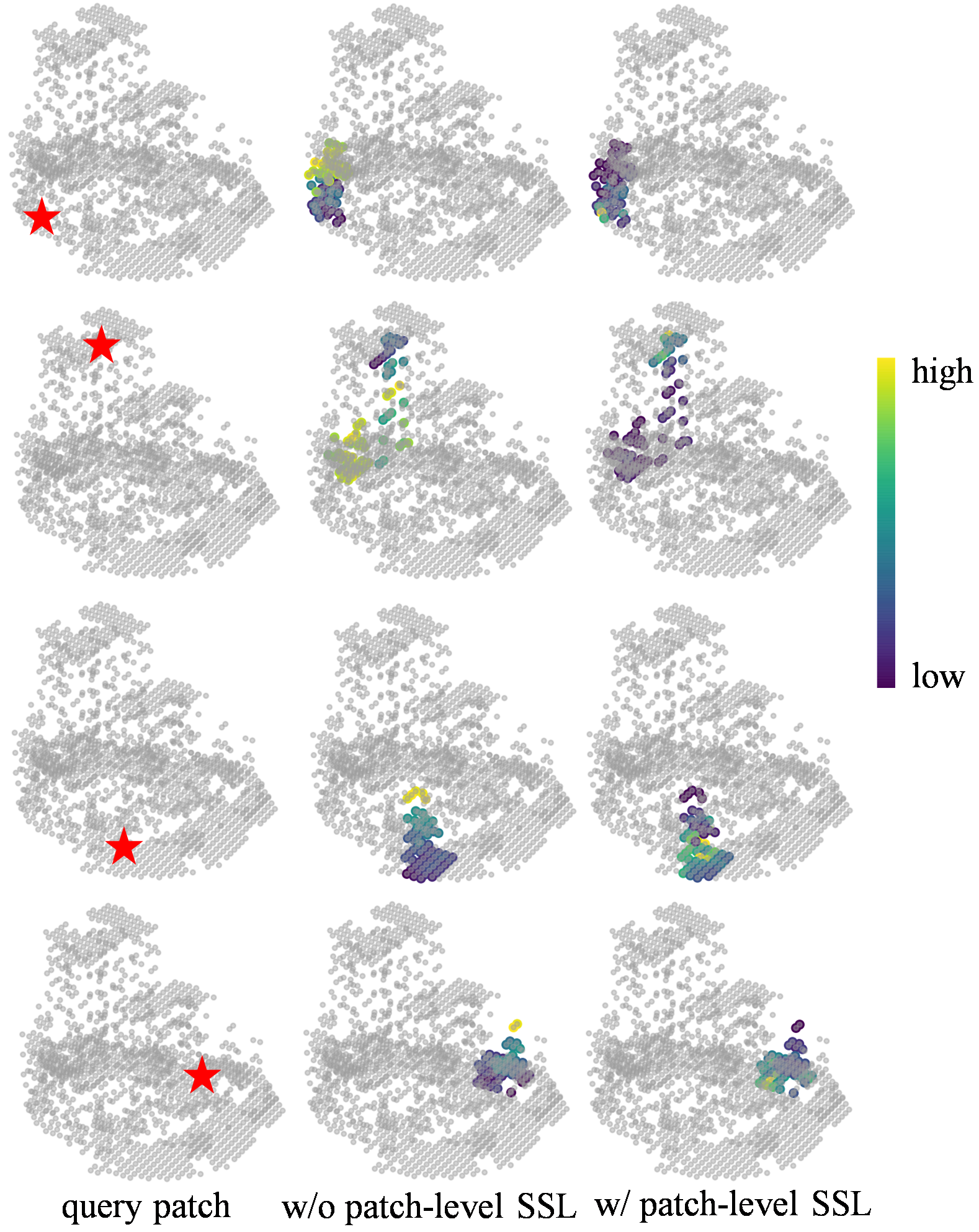}
\caption {Attention maps within octree windows.}
\label{fig_attn}
\end{figure}

\begin{figure}[] 
\centering
\includegraphics[width=0.9\linewidth, keepaspectratio]{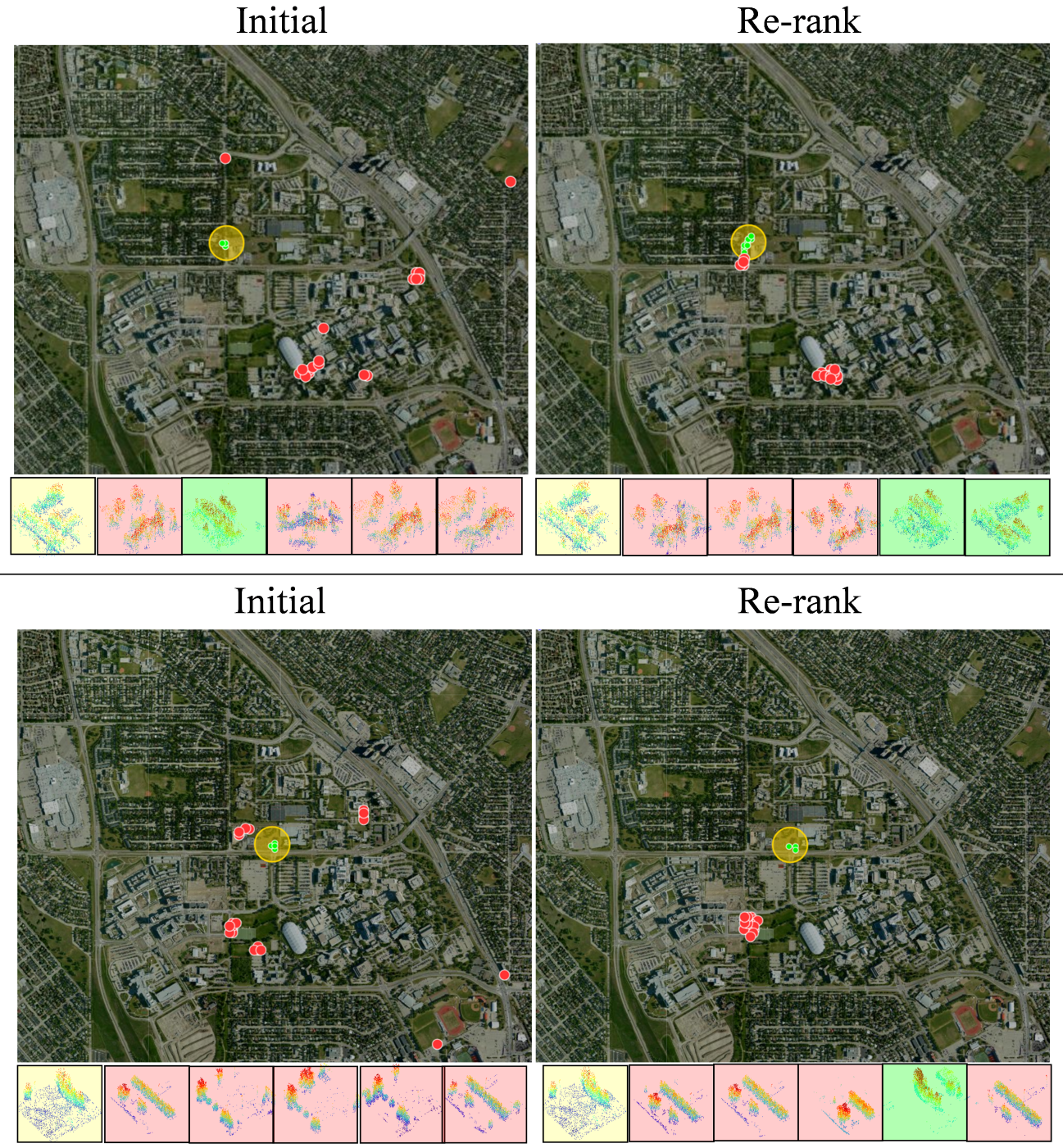}
\caption {Failure cases of re-ranking. Top: initial and re-ranked top-25 candidates. Bottom: the query and its top-5 candidates (from left to right). Positive and negative candidates are colored green and red, respectively.}
\label{fig_failure}
\end{figure}

\textbf{Efficiency.} We evaluate the runtime on the CS-Urban-Scenes dataset, which comprises 1,826 ground query scans and a database of 35,212 ALS point clouds. To evaluate the efficiency of different retrieval methods, we calculate the average inference time per query. As shown in Table~\ref{tab:retrieval_time}, compared to sparse convolution-based networks like MinkLoc3D and CrossLoc3D, Transformer-based architectures generally have larger parameter sizes and higher inference times. Meanwhile, we calculate the re-ranking on the whole dataset. As illustrated in Table~\ref{tab:reranking_time}, $k$-reciprocal is the slowest because it applies query expansion and computes Jaccard distances for all pairs. Our Expanded Reciprocal (ER) also builds reciprocal neighbors on the joint set, but we average features within each expanded neighborhood and score pairs from both sides. Consequently, it runs faster than $k$-reciprocal while remaining more thorough than SuperGlobal, which only considers top database candidates.

\textbf{Attention map.} To visualize the impact of our patch-level self-supervised learning (SSL), we visualize the attention maps within the octree partition windows. As illustrated in Figure~\ref{fig_attn}, we select four queries and plot the attention weights over their neighboring patches. Without patch-level SSL, the attention weights are scattered across non-essential patches. In contrast, when our patch-level SSL is utilized, the attention maps become more discriminative and focused on the semantic relations of the query patch.

\textbf{Deficiencies.} As shown in Figure~\ref{fig_failure}, we analyze the failure cases in both the retrieval and re-ranking stages. The first ground query contains a structured tree distribution. Although a true positive is retrieved at Top-2, the ER re-ranking fails due to ambiguities from other aerial submaps with similar tree layouts. The second query has large blank areas with sparse vegetation, which degrades feature representations and challenges both stages. These failures stem from the domain gap in different spatial coverage and noise variances, as well as scene challenges including repetitive or missing structures. Notably, even in failure cases, while the Top-25 retrieval candidates are scattered across the aerial map, our re-ranking method groups these candidates into several clusters. This behavior verifies our motivation of leveraging the structured spatial distribution of ALS point clouds.

\textbf{Future work.} Estimating accurate 6DoF poses remains an important direction for future work. Traditional methods typically retrieve candidates first and then employ fine registration for pose estimation \citep{yuan2024btc}. However, this coarse-to-fine localization paradigm easily fails when applied between aerial and ground point clouds. Due to the severe domain gap illustrated in Figure ~\ref{fig_difficulties}, establishing sufficient feature correspondences is challenging, making the feature matching process difficult. Furthermore, traditional registration approaches demand high memories, restricting efficient localization on city-scale ALS databases. Learning-based LiDAR global localization can be categorized into absolute pose regression (APR) \citep{li2024diffloc} and scene coordinate regression (SCR) \citep{wu2026leader}. APR methods directly regress global poses from input point clouds, whereas SCR methods estimate individual point or pixel coordinates. In addition to single-shot LiDAR global localization, some methods focus on sequential localization \citep{zou2025reliable}. However, most existing networks focus on ground-platform datasets, leaving aerial-ground LiDAR localization unexplored.

\section{Conclusion}
\label{sec_6}

We propose a robust aerial-ground LiDAR place recognition approach that bridges the domain gap through joint scene-level and patch-level self-supervised learning. To mitigate false positives from the retrieval results, our Expanded Reciprocal (ER) re-ranking algorithm leverages the structured spatial distribution of the aerial database and incorporates expanded reciprocal neighbors into the final distance metric. Evaluation of both retrieval and re-ranking stages on our newly collected CS-Urban-Scenes dataset shows that our pipeline outperforms current state-of-the-art methods. Future work will extend this framework to 6-DoF pose estimation in aerial-ground LiDAR global localization.

\printcredits

\section*{Declaration of interests}
The authors declare that they have no known competing financial interests or personal relationships that could have appeared to influence the work reported in this paper.

\bibliographystyle{cas-model2-names}

\bibliography{ref}

\end{sloppypar}
\end{document}